\documentclass{article}

\PassOptionsToPackage{numbers}{natbib}

    \usepackage[preprint,nonatbib]{neurips_2020}

\usepackage[utf8]{inputenc} %
\usepackage[T1]{fontenc}    %
\usepackage{hyperref}       %
\usepackage{url}            %
\usepackage{booktabs}       %
\usepackage{amsfonts}       %
\usepackage{nicefrac}       %
\usepackage{microtype}      %

\usepackage{comment}       %

\title{Belief Propagation Neural Networks}

\author{Jonathan Kuck\textsuperscript{1},\ \  Shuvam Chakraborty\textsuperscript{1},\ \ Hao Tang\textsuperscript{2},\ \ Rachel Luo\textsuperscript{1},\ \ Jiaming Song\textsuperscript{1},\\  \bf{Ashish Sabharwal}\textsuperscript{3},\ \ \normalfont{and}\ \ \bf{Stefano Ermon}\textsuperscript{1}\\
\textsuperscript{1}Stanford University\ \
\textsuperscript{2}Shanghai Jiao Tong University\ \
\textsuperscript{3}Allen Institute for Artificial Intelligence\\
{\tt\small \{kuck,shuvamc,rsluo,tsong,ermon\}@stanford.edu,} \\
{\tt\small silent56@sjtu.edu.cn, ashishs@allenai.org}
}

\usepackage[font={small}]{caption}

\usepackage{footmisc} %
\usepackage{booktabs} %
\usepackage{url}

\usepackage{natbib}

\renewcommand\cite{\citep}

\usepackage{mathtools}
\usepackage{amssymb}
\usepackage{multirow}
\usepackage{amsthm}

\usepackage{caption}
\usepackage{subcaption}

\newcommand{\iidsim}{\overset{\mathrm{i.i.d.}}{\sim}}

\newcommand{\beq}[1][\vspace{0.3em}]{#1\begin{equation}}
\newcommand{\eeq}{\end{equation}}

\newcommand{\bit}{\vspace{0mm}\begin{itemize}}
\newcommand{\eit}{\vspace{0mm}\end{itemize}}
\newcommand{\ben}{\vspace{0mm}\begin{enumerate}}
\newcommand{\een}{\vspace{0mm}\end{enumerate}}
\newtheorem{theorem}{Theorem}
\newtheorem{prop}{Proposition}

\newtheorem{lemma}{Lemma} 

\newtheorem{corollary}{Corollary}[theorem]
\newtheorem{definition}{Definition}

\newcommand{\LSE}[1]{\underset{#1}{\text{LSE }}}
\newcommand{\CONCAT}[2]{\overset{#2}{\underset{#1}{\text{CONCAT }}}}

\newcommand{\LNE}[0]{\text{LNE}}

\usepackage{xcolor}
\newcommand\jonathan[1]{\textcolor{orange}{[JK: #1]}}

\newcommand\se[1]{\textcolor{red}{[SE: #1]}}
\newcommand\ashish[1]{\textcolor{blue}{[AS: #1]}}
\newcommand\rachel[1]{\textcolor{green}{[RL: #1]}}
\newcommand\TD[1]{\textcolor{magenta}{[TD: #1]}}
 \renewcommand\jonathan[1]{} %
 \renewcommand\se[1]{} %
 \renewcommand\ashish[1]{} %
 \renewcommand\rachel[1]{} %
 \renewcommand\TD[1]{} %

\begin{document}

\maketitle

\begin{abstract}
Learned neural solvers have successfully been used to 
solve 
combinatorial 
 optimization and decision problems. More general counting variants of these problems, however, are still largely solved with hand-crafted solvers. 
 To bridge this gap, 
 we introduce belief propagation neural networks (BPNNs), a class of parameterized
 operators that operate on factor graphs and generalize Belief Propagation (BP).
In its strictest form, a BPNN layer (BPNN-D) is a learned iterative operator that provably maintains 
many of the desirable properties of BP for any choice of the parameters. 
Empirically, we show that by training BPNN-D learns to perform the task better than the original BP: it converges
1.7x faster 
on Ising models while providing tighter bounds.
On challenging model counting problems, BPNNs compute estimates 100's of times faster than state-of-the-art handcrafted methods, while returning an estimate of comparable quality.%
\end{abstract}

\section{Introduction}

Probabilistic inference problems arise in many domains, from statistical physics to machine learning. 
There is little hope that efficient, exact solutions to these problems exist as they are at least as hard as NP-complete decision problems.  Significant research has been devoted across the fields of machine learning, statistics, and statistical physics to develop variational and sampling based methods to approximate these challenging problems ~\cite{chandler1987introduction,mezard2002analytic,wainwright2008graphical,baxter2016exactly,mcbook}.  Variational methods such as
Belief Propagation (BP)~\cite{koller2009probabilistic} have been particularly successful at providing principled approximations due to extensive theoretical analysis.

We introduce belief propagation neural networks (BPNNs), a flexible neural architecture designed to estimate the partition function of a factor graph.  %
BPNNs generalize BP and can thus provide more accurate estimates than BP when trained on a small number of factor graphs with known partition functions.  At the same time, BPNNs retain many of BP's properties, which results in more accurate estimates compared to general neural architectures.
BPNNs are composed of iterative layers (BPNN-D) and an optional Bethe free energy layer (BPNN-B), both of which maintain the symmetries of BP under factor graph isomorphisms.  
BPNN-D is a parametrized iterative operator that strictly generalizes BP while preserving many of BP's guarantees.  Like BP, BPNN-D is guaranteed to converge on tree structured factor graphs and return the exact partition function.  For factor graphs with loops, BPNN-D computes a lower bound whenever the Bethe approximation obtained from fixed points of BP is a provable lower bound (with mild restrictions on BPNN-D).  BPNN-B performs regression from the trajectory of beliefs (over a fixed number of iterations) to the partition function of the input factor graph.  While this sacrifices some guarantees, the additional flexibility introduced by BPNN-B generally improves estimation performance.

Experimentally, we show that on Ising models BPNN-D is able to converge faster than standard BP and frequently finds better fixed points that provide tighter lower bounds.  BPNN-D generalizes well to Ising models sampled from a different distribution than seen during training and to models with nearly twice as many variables as seen during training, providing estimates of the log partition function that are significantly better than BP or a standard graph neural network (GNN) in these settings. We also perform experiments on community detection problems, where BP is known to perform well both empirically and theoretically, and show improvements over BP and a standard GNN. 
We then perform experiments on approximate model counting~\citep{Stockmeyer1983TheCO,Jerrum1986RandomGO,Karp1989MonteCarloAA,Bellare1998UniformGO}, the problem of computing the number of solutions to a Boolean satisfiability (SAT) problem.  Unlike the first two experiments it is very difficult for BP to converge in this setting.
Still, we find that BPNN learns to estimate accurate model counts from a training set of 10's of problems and generalize to problems that are significantly harder for an exact model counter to solve.
Compared to handcrafted approximate model counters, BP returns comparable estimates 100's times faster using GPU computation.

\section{Background: Factor Graphs and Belief Propagation}

In this section we provide background on factor graphs and belief propagation~\citep{koller2009probabilistic}.  A factor graph is a representation of a discrete probability distribution that takes advantage of independencies between variables to make the representation more compact.  Belief propagation is a method for approximating the normalization constant, or partition function, of a factor graph.
Let $p(\mathbf{x})$ be a discrete probability distribution defined in terms of a factor graph as
\begin{equation} \label{eq:distribution}
    p(\mathbf{x}) = \frac{1}{Z}  \prod_{a=1}^M f_a(\mathbf{x}_a) ,  \quad \ \ \ \ \ \ \   Z = \sum_{\mathbf{x}} \left ( \prod_{a=1}^M f_a(\mathbf{x}_a) \right ).
\end{equation}
where $\mathbf{x} = \{x_1, x_2, \dots, x_n\}$, $f_a(\mathbf{x}_a)>0$ are factors and $Z$ is the partition function. As a data structure, a factor graph is a bipartite graph with $n$ variables nodes and $M$ factor nodes.  
Factor nodes and variables nodes are connected if and only if the variable is in the scope of the factor.

\paragraph{Belief Propagation}

Belief propagation performs iterative message passing between neighboring variable and factor nodes.  Variable to factor messages, $m_{i \rightarrow a}^{(k)}(x_i)$, and factor to variable messages, $m_{a \rightarrow i}^{(k)}(x_i)$, are computed at every iteration $k$ as
\begin{equation} \label{eq:bp_msgs}
    m_{i \rightarrow a}^{(k)}(x_i) \coloneqq \prod_{c \in \mathcal{N}(i) \setminus a} m_{c \rightarrow i}^{(k-1)}(x_i),\,\, \text{and } m_{a \rightarrow i}^{(k)}(x_i) \coloneqq \sum_{\mathbf{x}_a \setminus x_i} f_a(\mathbf{x}_a) \prod_{j \in \mathcal{N}(a) \setminus i} m_{j \rightarrow a}^{(k)}(x_j).
\end{equation}

Messages are typically initialized either randomly or as constants. 
The BP algorithm estimates approximate marginal probabilities 
over the sets of variables $\mathbf{x}_a$ associated with each factor $f_a$.  We denote the belief over variables $\mathbf{x}_a$, after message passing iteration $k$ is complete,  as
 $b_a^{(k)}(\mathbf{x}_a) =  \frac{f_a(\mathbf{x}_a)}{z_a} \prod_{i \in \mathcal{N}(a)} m_{i \rightarrow a}^{(k)}(x_i)$
with $z_a = \sum_{\mathbf{x}_a} f_a(\mathbf{x}_a) \prod_{i \in \mathcal{N}(a)} m_{i \rightarrow a}^{(k)}(x_i)$.
Similarly, BP computes beliefs at each variable as $b_i^{(k)}(x_i) = \frac{1}{z_i} \prod_{a \in \mathcal{N}(i)} m_{a \rightarrow i}^{(k)}(x_i)$.
The belief propagation algorithm proceeds by iteratively updating variable to factor messages and factor to variable messages until they converge to fixed values, referred to as a fixed point of Equations~\ref{eq:bp_msgs}, or a predefined maximum number of iterations is reached.  
At this point the beliefs are used to compute a variational approximation of the factor graph's partition function.  This approximation, originally developed in statistical physics, is known as the Bethe free energy ${F_{\textrm{Bethe}} = U_{\textrm{Bethe}} - H_{\textrm{Bethe}}}\approx -\ln Z$~\cite{bethe1935statistical}.  It is defined in terms of the Bethe average energy $U_{\textrm{Bethe}} \coloneqq - \sum_{a=1}^M \sum_{\mathbf{x}_a} b_a(\mathbf{x}_a) \ln f_a(\mathbf{x}_a)$ and the Bethe entropy $H_{\textrm{Bethe}} \coloneqq - \sum_{a=1}^M \sum_{\mathbf{x}_a} b_a(\mathbf{x}_a) \ln   b_a(\mathbf{x}_a) + \sum_{i=1}^N (d_i - 1) \sum_{x_i} b_i(x_i) \ln b_i(x_i)$, where $d_i$ is the degree of variable node $i$.

\paragraph{Numerically Stable Belief Propagation.}
For numerical stability, belief propagation is generally performed in log-space and messages are normalized at every iteration.  It is also standard to add a \emph{damping} parameter, $\alpha \in [0,1)$, to improve convergence by taking partial update steps.  BP without damping is recovered when $\alpha=0$, while $\alpha=1$ would correspond to not updating messages and instead retaining their values from
the previous iteration.  With these modifications, 
the variable to factor messages from Equation~\ref{eq:bp_msgs} are rewritten as follows, where terms scaled by $\alpha$ represent the difference in the message's value from the previous iteration:
\begin{equation} \label{eq:varToFacMsgs_log}
     \overline{m}_{i \rightarrow a}^{(k)} = \tilde{m}_{i \rightarrow a}^{(k)} + \alpha  \big(\overline{m}_{i \rightarrow a}^{(k-1)} - \tilde{m}_{i \rightarrow a}^{(k)}\big),\,\, \text{where } \tilde{m}_{i \rightarrow a}^{(k)} = -z_{i \rightarrow a} + \sum_{c \in \mathcal{N}(i) \setminus a} \overline{m}_{c \rightarrow i}^{(k-1)}.
\end{equation}
Similarly, the factor to variable messages from Equation~\ref{eq:bp_msgs} are rewritten as
\begin{equation} \label{eq:facToVarMsgs_log}
    \overline{m}_{a \rightarrow i}^{(k)} = \tilde{m}_{a \rightarrow i}^{(k)} + \alpha \big(\overline{m}_{a \rightarrow i}^{(k-1)} - \tilde{m}_{a \rightarrow i}^{(k)}\big),\,\, \tilde{m}_{a \rightarrow i}^{(k)}= -z_{a \rightarrow i} + \LSE{\mathbf{x}_a \setminus x_i} \bigg( \phi_a(\mathbf{x}_a) + \sum_{j \in \mathcal{N}(a) \setminus i} \overline{m}_{j \rightarrow a}^{(k)} \bigg),
\end{equation}

Note that $\overline{m}_{i \rightarrow a}^{(k)}$ and $\overline{m}_{a \rightarrow i}^{(k)}$ are vectors of length $|X_i|$, $\phi_a(\mathbf{x}_a) = \ln \left( f_a \left( \mathbf{x}_a \right) \right)$ denotes log factors, $z_{i \rightarrow a}$ and $z_{a \rightarrow i}$ are normalization terms, and we use the shorthand $\LSE{}$ for the log-sum-exp function: $\LSE{\mathbf{x}_a \setminus x_i} \Big(\phi_a(\mathbf{x}_a)\Big) = \ln \left( \sum_{\mathbf{x}_a \setminus x_i} \exp\Big(\phi_a(\mathbf{x}_a)\Big) \right)$.

\section{Belief Propagation Neural Networks}

We design belief propagation neural networks (BPNNs) as a family of graph neural networks that operate on factor graphs.  %
Unlike standard graph neural networks (GNNs), BPNNs do not resend messages between nodes, a property taken from BP known as avoiding `double counting' the evidence.  This property guarantees that BPNN-D described below is exact on trees (Theorem~\ref{thm:BPNN-D_exact_LB}). 
BPNN-D is a strict generalization of BP (Proposition~\ref{prop:bpnn_subsume_lbp}), but is still guaranteed to give a lower bound to the partition function upon convergence for a class of factor graphs (Theorem~\ref{thm:BPNN-D_exact_LB}) by finding fixed points of BP (Theorem~\ref{thm:no_new_fixed_points}).  Like BP, BPNN preserves the symmetries inherent to factor 
graphs (Theorem~\ref{thm:BPNN_symmetry_short}).

BPNNs consist of two parts.  First, iterative BPNN layers output messages, analogous to standard BP.  These messages are used to compute beliefs using the same equations as for BP.  Second, the beliefs are passed into a Bethe free energy layer (BPNN-B) which generalizes the Bethe approximation by performing regression from beliefs to $Z$.  Alternatively, when the standard Bethe approximation is used in place of BPNN-B, BPNN provides many of BP's guarantees.

\paragraph{BPNN Iterative Layers}
BPNN iterative layers are flexible neural operators that can operate on beliefs or message in a variety of ways. Here, we focus on a specific variant, BPNN-D, due to its strong convergence properties, and we refer the reader to Appendix \ref{Appendix_BPNN} for information on other variants.
The BPNN iterative damping layer (BPNN-D) modifies factor-to-variable messages (Equation~\ref{eq:facToVarMsgs_log}) using the output of a learned operator 
$H:\mathbb{R}^{\sum_{i=1}^n d_i |X_i|} \to \mathbb{R}^{\sum_{i=1}^n d_i |X_i|}$ 
in place of the conventional damping term $\alpha \big(\overline{m}_{a \rightarrow i}^{(k-1)} - \tilde{m}_{a \rightarrow i}^{(k)}\big)$,  %
where $d_i$ denotes the degree and $|X_i|$ the cardinality of variable $X_i$. This learned operator $H(\cdot)$ takes as input the difference between iterations $k-1$ and $k$ of every factor-to-variable message, and modifies these differences jointly. It can thus be much richer than a scalar multiplier.  BPNN-D factor-to-variable messages are given by

\begin{equation} \label{eq:bpnn_varToFac_strong_guarantees}
    \overline{n}_{a \rightarrow i}^{(k)} = \tilde{n}_{a \rightarrow i}^{(k)} +  \Delta^{(k)}_{a \rightarrow i},\,\, \tilde{n}_{a \rightarrow i}^{(k)}= -z_{a \rightarrow i} + \LSE{\mathbf{x}_a \setminus x_i} \bigg( \phi_a(\mathbf{x}_a) + \sum_{j \in \mathcal{N}(a) \setminus i} \overline{n}_{j \rightarrow a}^{(k)} \bigg),
\end{equation}

where $\Delta^{(k)} = H\big(\overline{n}^{(k-1)} - \tilde{n}^{(k)}\big)$ denotes the result of applying $H(\cdot)$ to all factor-to-variable message differences and $\Delta^{(k)}_{a \rightarrow i}$ is the output corresponding to the modified $a \rightarrow i$ message difference.
Variable-to-factor messages are unchanged from Eq.~\ref{eq:varToFacMsgs_log}, except for taking messages $\overline{n}_{a \rightarrow i}^{(k)}$ as input,
\begin{equation}\label{eq:bpnn_facToVar_strong_guarantees}
     n_{i \rightarrow a}^{(k)} = \tilde{n}_{i \rightarrow a}^{(k)} + \alpha (n_{i \rightarrow a}^{(k-1)} - \tilde{n}_{i \rightarrow a}^{(k)}),\,\, \text{where } \tilde{n}_{i \rightarrow a}^{(k)} = -z_{i \rightarrow a} + \sum_{c \in \mathcal{N}(i) \setminus a} \overline{n}_{c \rightarrow i}^{(k-1)}.
\end{equation}

Note that we recover Equations~\ref{eq:varToFacMsgs_log} and \ref{eq:facToVarMsgs_log} exactly if 
$H$ is an elementwise function $H(x) = \alpha x$. Thus:
\begin{prop}\label{prop:bpnn_subsume_lbp}
BPNN-Ds subsume BP and damped BP as a strict generalization.
\end{prop}

For non-trivial choices of $H(\cdot)$, whether BPNN preserves the fixed points of BP or introduces any new ones turns out to depend only on the set of fixed points of $H(\cdot)$ itself, i.e., $\{x \mid H(x) = x\}$. As we show next, this property allows us to easily enforce that every fixed point of BP is also a fixed point of BPNN-D (Theorem~\ref{thm:preserve_BP_fixed_points}), or vice versa (Theorem~\ref{thm:no_new_fixed_points}).\footnote{For lack of space, all proofs are deferred to Appendix~\ref{appendix:proofs}.}

\begin{theorem}\label{thm:preserve_BP_fixed_points}
If zero is a fixed point of 
$H(\cdot)$,
then every fixed point of BP is also a fixed point of BPNN-D.
\end{theorem}

\begin{theorem}\label{thm:no_new_fixed_points}
If 
$H(\cdot)$ does not have any non-zero fixed points,
then every fixed point of BPNN-D 
is also a fixed point of BP.
\end{theorem}

Combining Theorems~\ref{thm:preserve_BP_fixed_points} and \ref{thm:no_new_fixed_points}, we obtain Corollary~\ref{cor:identical_fixed_points}.  

\begin{corollary}\label{cor:identical_fixed_points}
If zero is the unique fixed point of 
$H(\cdot)$,
then the fixed points of BP and BPNN-D are identical. This property is satisfied when $H(x) = x + \bar{H}(x) - \bar{H}(\mathbf{0})$ for any invertible function $\bar{H}(\cdot)$. %
\end{corollary}

Note that a broad class of highly expressive learnable operators are invertible \cite{behrmann2019invertible}. %
Enforcing that every fixed point of BPNN-D is also a fixed point of BP is particularly useful, as it immediately follows that BPNN-D returns a lower bound whenever the Bethe approximation obtained from fixed points of BP returns a provable lower bound (Theorem~\ref{thm:BPNN-D_exact_LB}). When a BPNN-D layer is applied iteratively until convergence, fast convergence is guaranteed for tree structured factor graphs (Proposition~\ref{prop:bpnn_convergence}). As mentioned, BPNN iterative layers are flexible and can additionally be modified to operate directly on message values or factor beliefs %
at the expense of no longer returning a lower bound (see Appendix \ref{Appendix_BPNN}).

\begin{theorem}\label{thm:BPNN-D_exact_LB}
If zero is the unique fixed point of 
$H(\cdot)$,
the Bethe approximation computed from beliefs at a fixed point of BPNN-D (1) is exact for tree structured 
graphs and (2) lower bounds the partition function of any factor graph with binary variables and log-supermodular potential functions.
\end{theorem}

\begin{prop}
\label{prop:bpnn_convergence}
BPNN-D converges within $\ell$ iterations on tree structured factor graphs with height $\ell$.
\end{prop}

\paragraph{Bethe Free Energy Layer (BPNN-B).} %
When convergence to a fixed point is unnecessary, we can increase the flexibility of our architecture by building a K-layer BPNN from iterative layers that do not share weights. Additionally we define a Bethe free energy layer (BPNN-B, Equation~\ref{eq:bethe_MLP}) using two MLPs that take the trajectories of learned beliefs from each factor and variable as input and output scalars:
\begin{align}\label{eq:bethe_MLP}
\begin{split}
      & f_{\textrm{BPNN}}(G_{\textrm{factor}}) =  \sum_{i=1}^n \text{MLP}_{BV} \Bigg[ \CONCAT{k=1}{K}\bigg( (d_i - 1)  b_i^{(k)}(x_i) \ln b_i^{(k)}(x_i) \bigg)  \Bigg] + \\
    &      \frac{1}{\lvert \mathbf{x}_a \rvert !}\sum_{a=1}^{M} \sum_{\sigma \in S_{\lvert \mathbf{x}_a \rvert}} \text{MLP}_{BF} \Bigg[    \CONCAT{k=1}{K} \Bigg(
    \sigma \Big(b_a^{(k)}(\mathbf{x}_a) \ln f_a(\mathbf{x}_a)\Big),  
     \sigma\Big(-b_a^{(k)}(\mathbf{x}_a) \ln   b_a^{(k)}(\mathbf{x}_a) \Big) \Bigg) \Bigg].\\
\end{split}
\end{align}

This parameterization subsumes the standard Bethe approximation, so we can initialize the parameters of $f_{\textrm{BPNN}}$ to output the Bethe approximation computed from the final layer beliefs (see the appendix for details). Note that $\lvert \mathbf{x}_a \rvert$ is the number of variables in the scope of factor $a$, $S_{\lvert \mathbf{x}_a \rvert}$ denotes the symmetric group (all permutations of $\{1,2,\dots,\lvert \mathbf{x}_a \rvert\}$), and the permutation $\sigma$ is applied to the dimensions of all $2k$ concatenated terms.  We ensure that BPNN preserves the symmetries of BP (Theorem~\ref{thm:BPNN_symmetry_short}) by passing all factor permutations through $\text{MLP}_{BF}$ and averaging the result.  %

\paragraph{BPNN Preserves the Symmetries of BP.}
BPNN is designed so that equivalent input factor graphs are mapped to equivalent outputs.  This is a property that BP satisfies by default.  Standard GNNs are also designed to satisfy this property, however the notion of `equivalence' between graphs is different than `equivalence' between factor graphs.  In this section we formalize these statements.

Graph isomorphism defines an equivalence relationship between graphs that is respected by standard GNNs.  Two isomorphic graphs are structurally equivalent and indistinguishable if the nodes are appropriately matched.  More formally, there exists a bijection between nodes (or their indices) in the two graphs that defines this matching.  Standard GNNs are designed so that output node representations are equivariant to the input node indexing; the indexing of output node representations matches the indexing of input nodes.  Output node representations of a GNN run on two isomorphic graphs can be matched using the same bijection that defines the isomorphism. Further, standard GNNs are designed to map isomorphic graphs to the same graph-level output representation.  These two properties are achieved by using a message aggregation function and a graph-level output function that are both invariant to node indexing.

We formally define factor graph isomorphism in Definition~\ref{def:fg_isomorphism} (Appendix~\ref{appendix:proofs}).  This equivalence relationship is more complicated than for standard graphs because factor potentials define a structured relationship between factor and variable nodes. %
As in a standard graph, variable nodes are indexed globally ($X_1, X_2, \dots, X_n$) in the representation of a factor graph.  Additionally, variable nodes are also indexed locally by factors that contain them.  %
This is required because each factor dimension (note that factors are tensors) corresponds to a unique variable, unless the factor happens to be symmetric. %
Local variable indices define a mapping between factor dimensions and the variables' global indices.  These local variable indices lead to additional bijections in the definition of isomorphic factor graphs (condition 2 in Definition~\ref{def:fg_isomorphism}).  Note that standard GNNs do not respect factor graph isomorphisms because of these additional bijections. 

In contrast to standard GNNs, BP respects factor graph isomorphisms.  When BP is run on two isomorphic factor graphs for the same number of iterations with constant message initialization\footnote{\label{note1}Any message initialization can be used, as long as initial messages are equivariant, see Lemma~\ref{lemma:msg_equiv}.} %
the output beliefs and messages satisfy bijections corresponding to those of the input factor graphs.  Specifically, messages are equivariant to global node indexing (Lemma~\ref{lemma:msg_equiv}), variable beliefs are equivariant to global variable node indexing (Lemma~\ref{lemma:var_bel_equiv}), and factor beliefs are equivariant to global factor node indexing and local variable node indexing within factors (Lemma~\ref{lemma:fac_bel_equiv}). 
We refer to the above properties as \emph{equivariances of BP} under factor graph isomorphisms. We show that these properties also apply to BPNN-D when $H(\cdot)$ is equivariant to global node %
indexing. The Bethe approximation obtained from isomorphic factor graphs is identical, when BP is run for the same number of iterations with constant message initialization\footref{note1}. %
BPNN-B also satisfies this property because it is, by design, invariant to local variable indexing within factors (Lemma~\ref{lemma:BPNN-B_invariance}).  Together, these properties lead to the following:

\begin{theorem}\label{thm:BPNN_symmetry_short}
If $H(\cdot)$ is equivariant to global node indexing, then (1) BPNN-D messages and beliefs preserve the equivariances of BP under factor graph isomorphisms and (2) BPNN-B is invariant under factor graph isomorphisms.
\end{theorem}

\section{Experiments}
In our experiments we trained BPNN to estimate the partition function of factor graphs from a variety of domains.  First, experiments on synthetic Ising models show that BPNN-D can learn to find better fixed points than BP and converge faster.  Additionally, BPNN generalizes to Ising models with nearly twice as many variables as those seen during training and  that were sampled from a different distribution.  Second, experiments and an ablation study on the stochastic block model from community detection show that maintaining properties of BP in BPNN improves results over standard GNNs.  Finally, model counting experiments performed on real world SAT problems show that BPNN can learn from 10's of training problems, generalize to problems that are harder for an exact model counter, and compute estimates 100's of times faster than handcrafted approximate model counters.  We implemented our BPNN and the baseline GNN using PyTorch Geometric~\citep{Fey/Lenssen/2019}. We refer the reader to Appendix \ref{Appendix_GNN} for details on the GNN. %
\subsection{Ising Models}

\begin{figure*}
     \centering
     \begin{subfigure}[b]{0.49\textwidth}
         \centering
         \includegraphics[width=\textwidth]{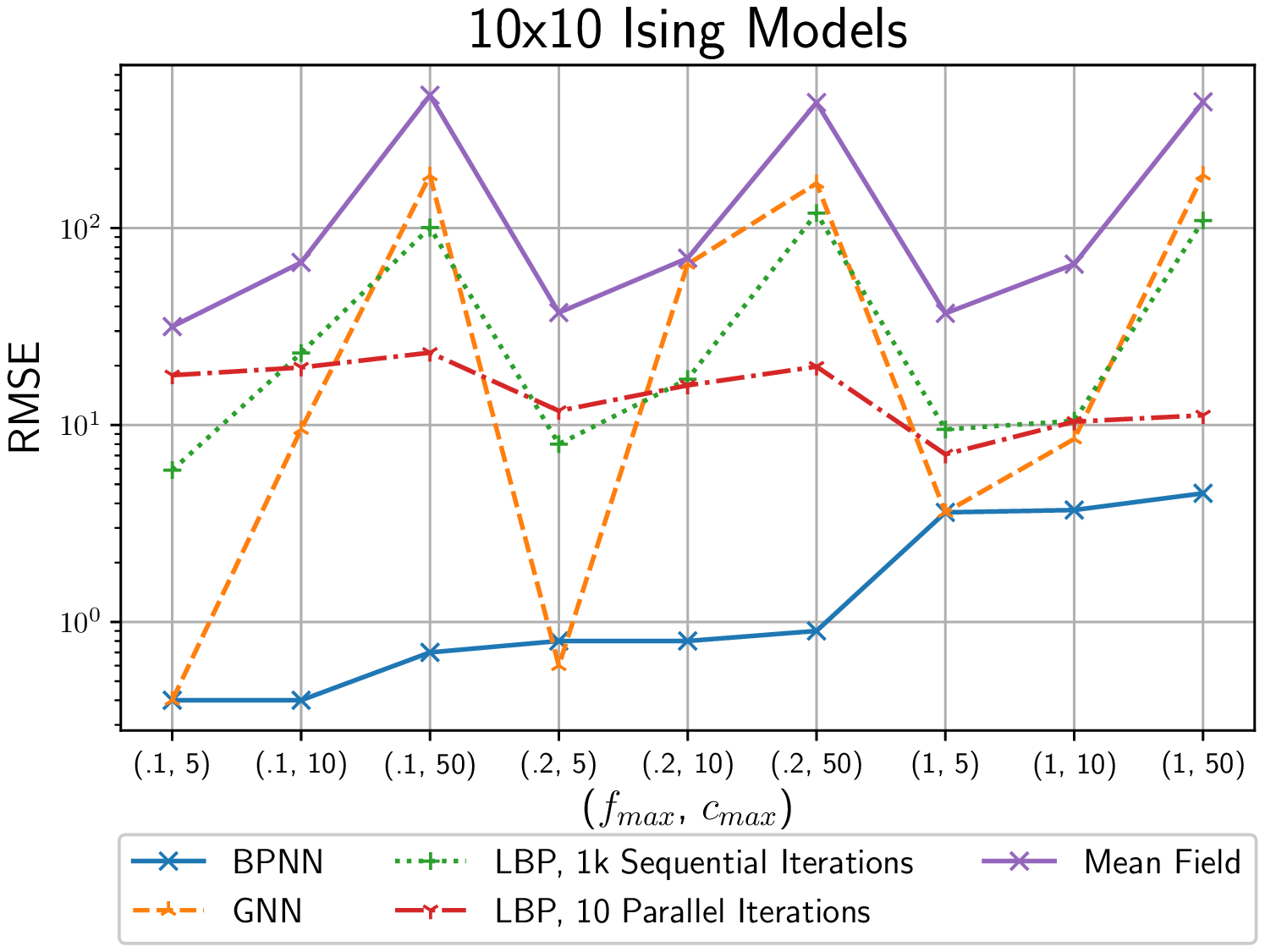}
     \end{subfigure}
     \hfill
     \begin{subfigure}[b]{0.49\textwidth}
         \centering
         \includegraphics[width=\textwidth]{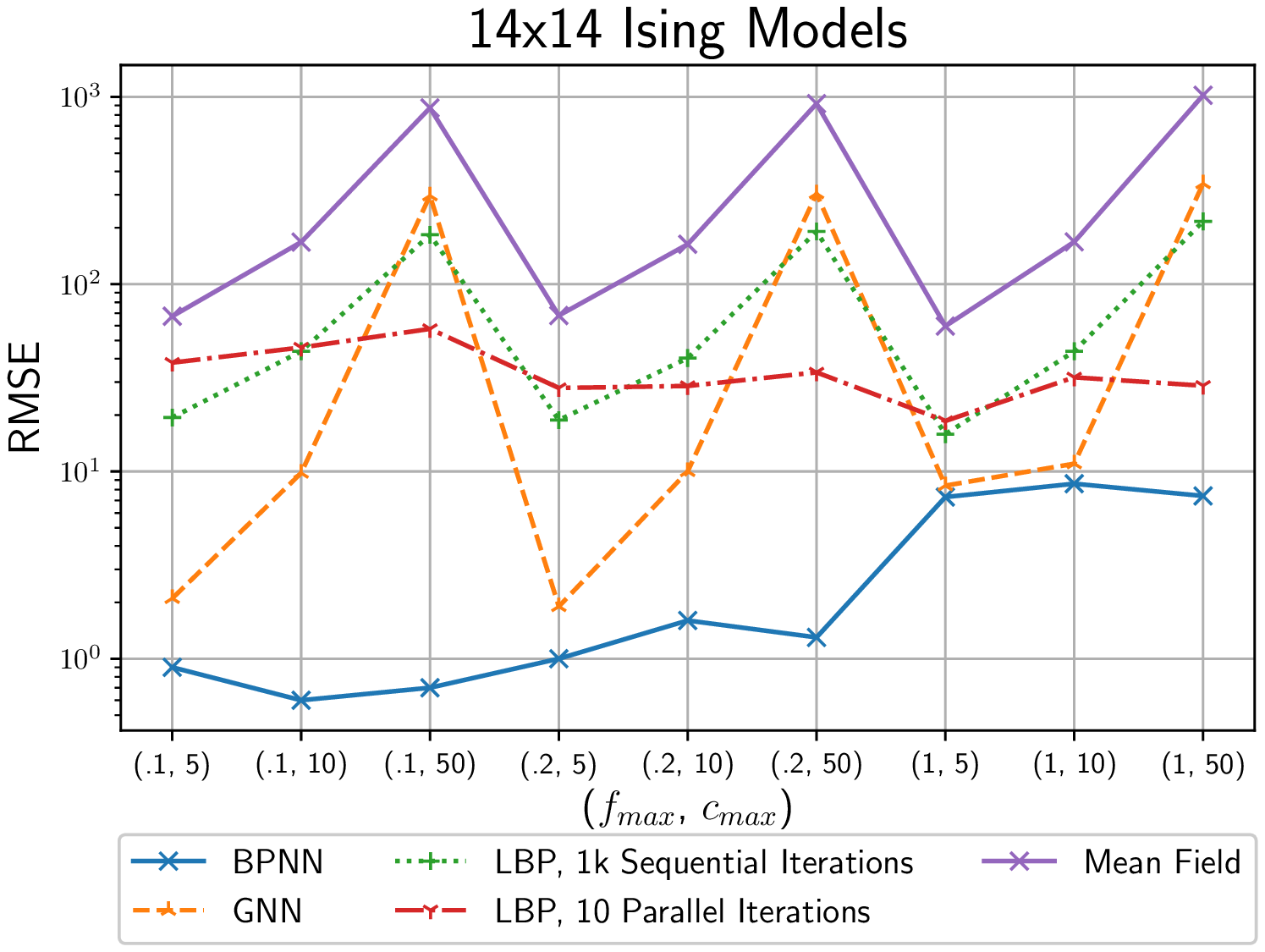}
     \end{subfigure}
        \caption{Each point represents the root mean squared error (RMSE, y-axis) of the specified method on a test set of 50 Ising models sampled with the parameters $f_\text{max}$ and $c_\text{max}$ (x-axis). The leftmost point shows results for test data drawn from the same distribution as training. BPNN significantly improves upon loopy belief propagation (LBP) for both in and out of distribution data. BPNN also significantly outperforms GNN on out of distribution data and larger models.  }
        \label{fig:ising_model_OOD}
\end{figure*}

We followed a common experimental setup used to evaluate approximate integration methods~\cite{hazan2012partition,ermon2013taming}.  We randomly generated grid structured attractive Ising models whose partition functions can be computed exactly using the junction tree algorithm~\cite{lauritzen1988local} for training and validation.  BP computes a provable lower bound for these Ising models~\cite{Ruozzi2012TheBP}.  This family of Ising models is only slightly more general than the one studied in \citep{Koehler2019FastCO}, where BP was proven to quickly converge to the Bethe free energy's global optimum. We found that an iterative BPNN-D layer was able to converge faster than standard BP and could find tighter lower bounds for these problems.  Additionally we trained a 10 layer BPNN and evaluated its performance against a 10 layer GNN architecture (details in Appendix). 
Compared to the GNN, BPNN has improved generalization when tested on larger Ising models and Ising models sampled from a different distribution than seen during training.%

\paragraph{Improved Lower Bounds and Faster Convergence.}
We trained an iterative BPNN-D layer to lower bound the partition function on a training set of 50 random Ising models of size 10x10 (100 variables). (See the appendix for further details.)  We then ran the learned BPNN-D and standard BP on a validation set of 50 Ising models. We empirically verified that BPNN-D found fixed points corresponding to tighter lower bounds than BP, and that it found them faster than standard BP.  BPNN-D converged on all 50 models, while BP failed to converge within 200 iterations for 6 of the models. We recorded the number of iterations that BPNN-D and BP run with parallel updates took to converge, defined as a maximum factor-to-variable message difference of $10^{-5}$. BPNN-D had a median improvement ratio of 1.7x over BP, please refer to the appendix for complete convergence plots.  Among the 44 models where BP converged, the RMSE between the exact log partition function and BPNN-D's estimate was .97 compared with 7.20 for BP.  For 10 of the 44 models, BPNN-D found fixed points corresponding to lower bounds on the log partition function that were larger (i.e., better) than BP's by 3 to 22 (corresponding to bounds on the partition function that were 20 to $e^{22}$ times larger).  In contrast, the log lower bound found by BP was never larger than the bound found by BPNN-D by more than 1.7.

\paragraph{Out of Distribution Generalization.}
We tested BPNN's ability to generalize to larger factor graphs and to shifts in the test distribution.  Again we used a training set of 50 Ising models of size 10x10 (100 variables).
We sampled test Ising models from distributions with generative parameters increased by factors of 2 and 10 from their training values (see appendix for details) and with their size increase to 14x14 (for 196 variables instead of the 100 seen during training). For this experiment we used a BPNN architecture with 10 iterative layers whose weights were not tied and with MLPs that operate on factor messages (without a BPNN-B layer).  As a baseline we trained a 10 layer GNN (maximally powerful GIN architecture) with width 4 on the same dataset.  We also compute the Bethe approximation from running standard loopy belief propagation and the mean field approximation.  We used the libDAI~\cite{Mooij_libDAI_10} implementation for both.  We tested loopy belief propagation with and without damping and with both parallel and sequential message update strategies.  We show results for two settings whose estimates of the partition function differ most drastically: (1) run for a maximum of 10 iterations with parallel updates and damping set to .5, and (2) run for a maximum of 1000 iterations with sequential updates using a random sequence and no damping. 
Full test results are shown in Figure~\ref{fig:ising_model_OOD}.  The leftmost point in the left figure shows results for test data that was drawn from the same distribution used for training the BPNN and GNN.  The BPNN and GNN perform similarly for data drawn from the same distribution seen during training.  However, our BPNN significantly outperforms the GNN when the test distribution differs from the training distribution and when generalizing to the larger models.  Our BPNN also significantly outperforms loopy belief propagation, both for test data drawn from the training distribution and for out of distribution data.  %

\subsection{Stochastic Block Model}
The Stochastic Block Model (SBM) is a generative model describing the formation of communities and is often used to benchmark community detection algorithms~\cite{JMLR:v18:16-480}.  While BP does not lower bound the partition functions of associated factor graphs for SBMs, it has been shown that BP asymptotically (in the number of nodes) reaches the information theoretic threshold for community recovery on SBMs with fewer than 4 communities ~\cite{JMLR:v18:16-480}.
We trained a BPNN to estimate the partition function of the associated factor graph and observed improvements over estimates obtained by BP or a maximally powerful GNN, which lead to more accurate marginals that can be used to better quantify uncertainty in SBM community membership.
We refer the reader to Appendix \ref{SBM Appendix} for a formal definition of SBMs as well as our procedure for constructing factor graphs from a sampled SBM.

\paragraph{Dataset and Methods}
In our experiments, we consider SBMs with 2 classes and 15-20 nodes, so that exact inference is possible using the Junction Tree algorithm. In this non-asymptotic setting, BP is a strong baseline and can almost perfectly recover communities~\citep{Chen2019SupervisedCD}, but is not optimal and thus does not compute exact marginals or partition functions.
For training, we sample 10 two class SBMs with 15 nodes, class probabilities of .75 and .25, and edge probability of .93 within and .067 between classes along with four such graphs for validation. For each graph, we fix each node to each class and calculate the exact log partition using the Junction Tree Algorithm, producing 300 training and 120 validation graphs. We explain in Appendix \ref{SBM Appendix} how these graphs can be used to calculate marginals.

To estimate SBM partition functions, we trained a BPNN with 30 iterative BPNN layers that operate on messages (see Appendix \ref{Appendix_BPNN}), followed by a BPNN-B layer. Since BP does not provide a lower bound for SBM partitions, we took advantage of BPNN's flexibility and chose greater expressive power over BPNN-D's superior convergence properties. We compared against BP 
and a GNN as baseline methods. Additionally, we performed 2 ablation experiments. We trained a BPNN with a BPNN-B layer that was not permutation invariant to local variable indexing, by removing the sum over permutations in $S_{\lvert \mathbf{x}_a \rvert}$ from Equation~\ref{eq:bethe_MLP} and only passing in the original beliefs.  We refer to this non-invariant version as BPNN-NI. We then forced BPNN-NI to `double count' messages by changing the sums in Equations~\ref{eq:bpnn_varToFac_strong_guarantees} and \ref{eq:bpnn_facToVar_strong_guarantees} to be over $j \in \mathcal{N}(a)$.  We refer to this non-invariant version that performs double counting as BPNN-DC. We refer the reader to Appendix \ref{SBM Appendix} for further details on models and training.

\paragraph{Results}
As shown in Table \ref{tab:bpnn_sbm_trainingval_15}, BPNN provides the best estimates for the partition function. Critically, we see that not 'double counting' messages and preserving the symmetries of BP are key improvements of BPNN over GNN. Additionally, BPNN outperforms BP and GNN on out of distribution data and larger graphs and can learn more accurate marginals. %
We refer the reader to Appendix \ref{SBM Appendix} for more details on these additional experiments.

\begin{table}
    \begin{center}
    \centering
    \begin{tabular}{c c c c c}
    \toprule
    \multicolumn{5}{c}{\textbf{Stochastic Block Model RMSE}}  \\
    BP & GNN & BPNN-DC & BPNN-NI & BPNN \\
    Train/Val & Train/Val & Train/Val & Train/Val & Train/Val \\
    \hline
    12.55/11.14 & 7.33/7.93 & 7.04/8.43 & 4.43/5.63 & \textbf{4.16}/\textbf{4.15}\\ 
    \hline
    \end{tabular}
    \end{center}
    \caption{RMSE of SBM $\ln(Z)$ estimates. BPNN outperforms BP, GNN, and ablated versions of BPNN.}
    \label{tab:bpnn_sbm_trainingval_15}
\end{table}

\subsection{Model Counting}

In this section we use a BPNN to estimate the number of satisfy solutions to a Boolean formula, a challenging problem for BP which generally fails to converge due to the complex logical constraints and 0 probability states. %
Computing the exact number of satisfy solutions (exact model counting) is a \#P-complete problem~\cite{valiant1979complexity_SAT}.  Model counting is a fundamental problem that arises in many domains including probabilistic reasoning~\cite{Roth1993OnTH,belle2015hashing}, network reliability~\cite{DueasOsorio2017CountingBasedRE}, and detecting private information leakage from programs~\cite{Biondi2018ScalableAO}. However, the computational complexity of exact model counting has led to a significant body of work on approximate model counting~\citep{Stockmeyer1983TheCO,Jerrum1986RandomGO,Karp1989MonteCarloAA,Bellare1998UniformGO,mbound,ermon2014low,ivrii2015computing,achlioptasstochastic,achlioptas2018fast,SM19}, with the goal of estimating the number of satisfying solutions at a lower computational cost.  %

\paragraph{Training Setup.}
All BPNNs trained in this section were composed of 5 BPNN-D layers followed by a BPNN-B layer and were trained to predict the natural logarithm of the number of satisfying solutions to an input formula in CNF form.  This is accomplished by converting the CNF formula into a factor graph whose partition function is the number of satisfying solutions to the input formula. We evaluated the performance of our BPNN using benchmarks from \citep{SM19}, with ground truth model counts obtained 
using DSharp~\cite{Muise2012}.  The benchmarks fall into 7 categories, including network QMR problems (Quick Medical Reference)~\cite{jaakkola1999variational}, 
network grid problems, and bit-blasted versions of satisfiability modulo theories library (SMTLIB) benchmarks~\cite{CMV16}.  Each category contains 14 to 105 problems allocated for training and validation.  %
See the appendix for additional details on training, the dataset, and our use of minimal independent support variable sets.

\paragraph{Baseline Approximate Model Counters.}
For comparison we ran two state-of-the-art approximate model counters on all benchmarks, ApproxMC3
~\citep{CMV16,SM19} and F2
~\citep{achlioptas2017probabilistic,achlioptas2018fast}.  ApproxMC3 is a randomized hashing algorithm that returns an estimate of the model count that is guaranteed to be within a multiplicative factor of the exact model count with high probability.  F2 gives up the probabilistic guarantee that the returned estimate will be within a multiplicative factor of the true model count in return for significantly increased computational efficiency.
We also attempted to train a GNN, using the architecture from~\citep{selsam2018learning} adapted from classification to regression.  We used the author's code, slightly modified to perform regression, but were not successful in achieving non-trivial learning.

\begin{figure*}
     \centering
     \begin{subfigure}[b]{0.49\textwidth}
         \centering
         \includegraphics[width=\textwidth]{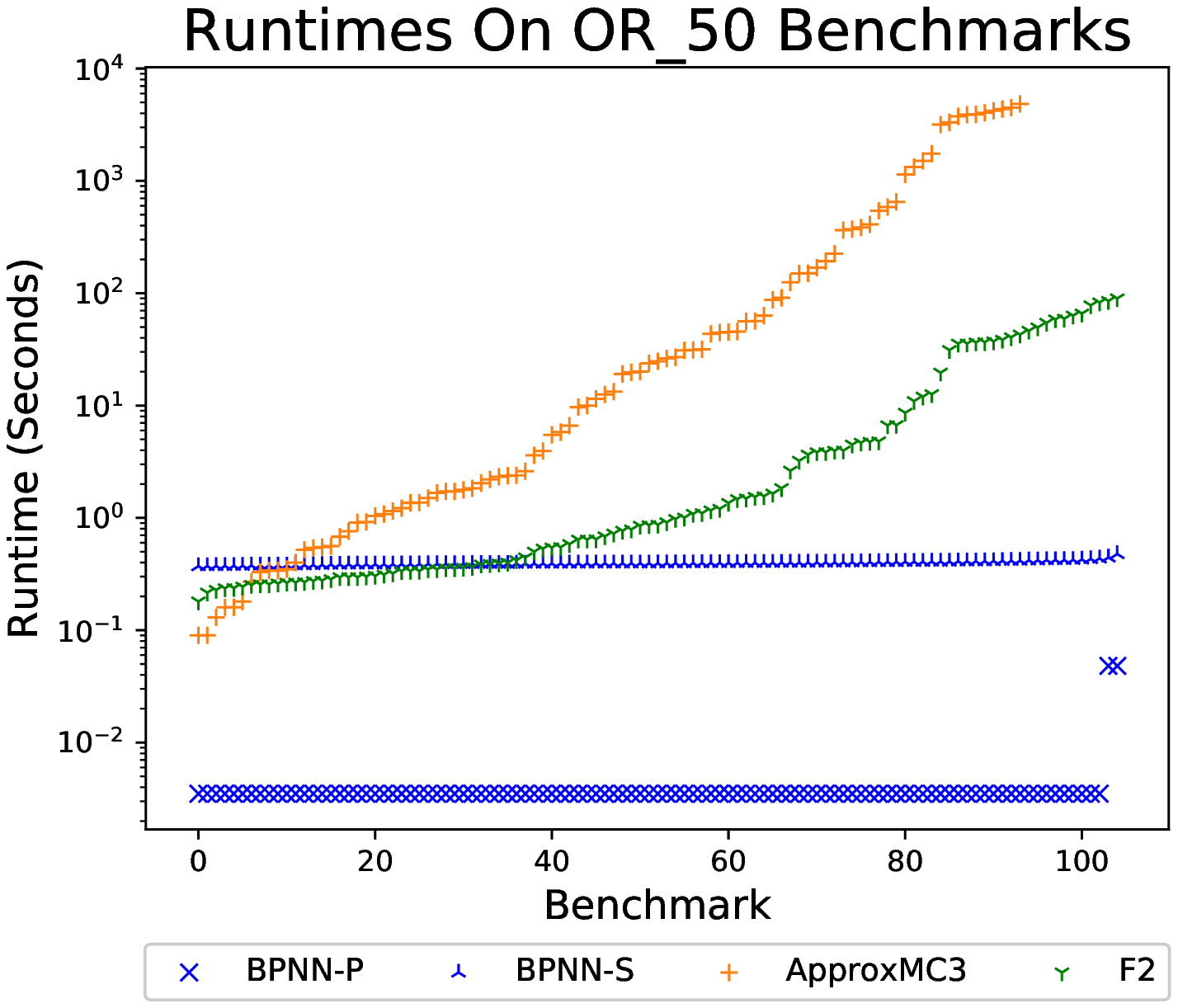}
     \end{subfigure}
     \hfill
     \begin{subfigure}[b]{0.49\textwidth}
         \centering
         \includegraphics[width=\textwidth]{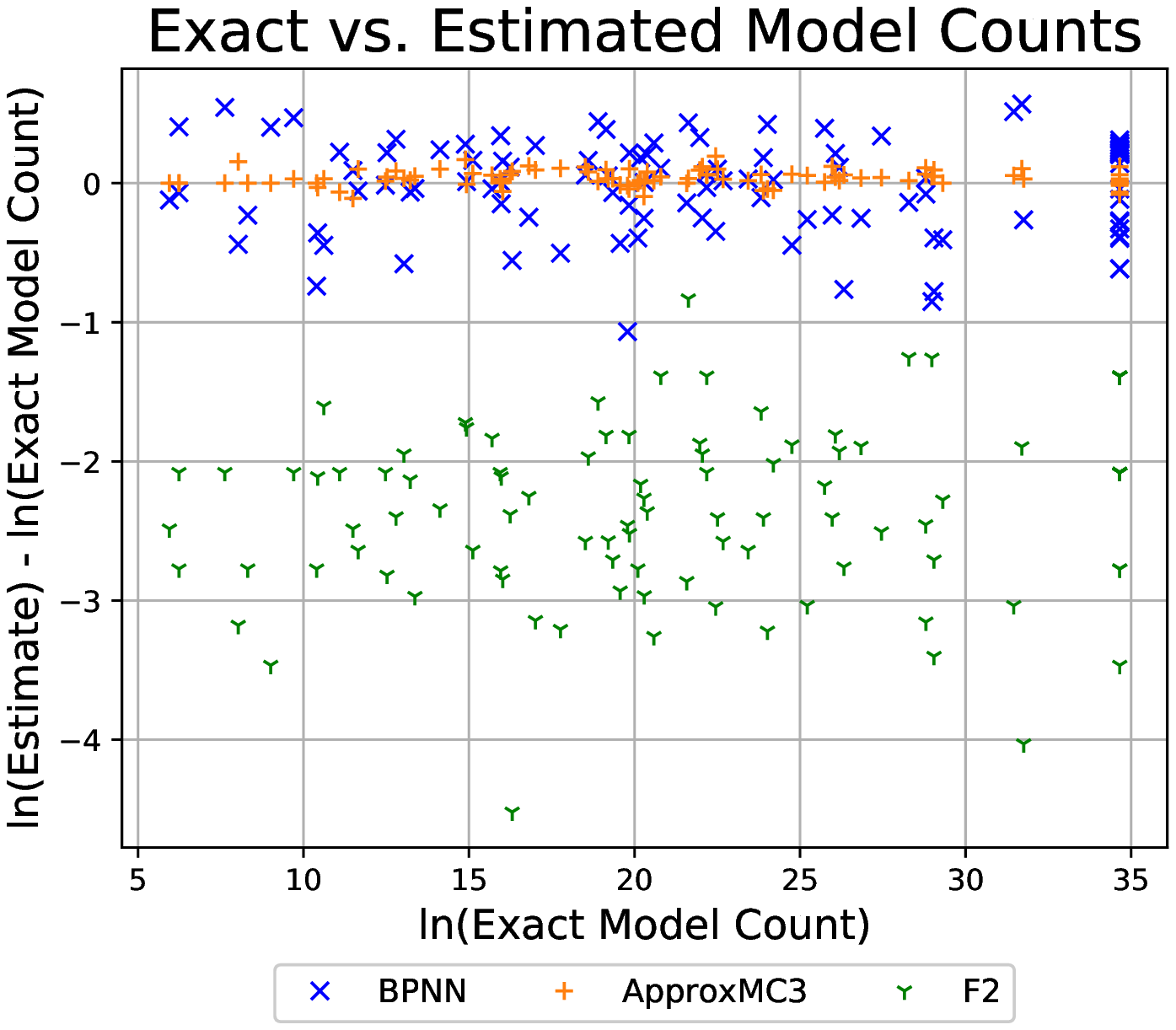}
     \end{subfigure}
        \caption{Left: cactus plot of runtimes for the 105 instances in the 'or\_50' category solved by BPNN, F2, and ApproxMC3.  BPNN-P denotes the time taken to run BPNN in parallel on a GPU divided by the number of instances per batch (batch size=103).  Median speedups of BPNN-P over F2 and ApproxMC among the plotted benchmarks are 248 and 3,689 respectively.  BPNN-S denotes the time taken to run BPNN sequentially on each instance (using a CPU).  Median speedups of BPNN-S over F2 and ApproxMC among the plotted benchmarks are 2.2 and 32, resp.  While BPNN solved each instance within 1 second, ApproxMC3 \emph{timed out} on 12 instances (out of 105) after 5000 seconds, which are not plotted.  Right: error in estimated log model count (base e) plotted against the exact model count for `or\_50' training and validation benchmarks.  BPNN's validation RMSE was .30 on this category compared with a RMSE of 2.5 for F2. 
        }
        \label{fig:bpnn_vs_F2_or50}     
\end{figure*}

\paragraph{BPNNs Provide Excellent Computational Efficiency.}
Figure~\ref{fig:bpnn_vs_F2_or50} shows runtimes and estimates for BPNN, ApproxMC3, and F2 on all benchmarks from the category `or\_50'.  BPNN is signficantly faster than both F2 and ApproxMC.  BPNN provides median speedups of 2.2 and 32 over F2 and ApproxMC3 when all methods are run using a CPU.  When BPNN is allowed to run in parallel on a GPU, it provides median speedups of 248 and 3,689 over F2 and and ApproxMC3.  Additionally, BPNN's estimates are significantly tighter than F2's, with a RMSE for BPNN of .30 compared with 2.5 for F2.   Please see the appendix for further runtime comparisons between methods.

\paragraph{Learning from Limited Data.}
We trained a separate BPNN on a random sampling of 70\% of the problems in each training category.  This gave each BPNN only 9 to 73 benchmarks to learn from.  In contrast, prior work has performed approximate model counting on Boolean formulas in disjunctive normal form (DNF) by creating a large training set of 100k examples whose model counts can be approximated with an efficient polynomial time algorithm~\cite{abboud2020learning}.  Such an algorithm does not exist for model counting on CNF formulas, making this approach intractable.  Nonetheless, BPNN achieves training and validation RMSE comparable to or better than F2 across the range of benchmark categories (see the appendix for complete results).  This demonstrates that BPNNs can capture the distribution of diverse families of SAT problems in an extremely data limited regime.

\paragraph{Generalizing from Easy Data to Hard Data.}
We repeated the same experiment from the previous paragraph, but trained each BPNN on the 70\% of the problems from each category that DSharp solved fastest.  Validation was performed on the remaining 30\% of problems that took longest for DSharp to solve.  These hard validation sets are significantly more challenging for Dsharp.  The median runtime in each category's hard validation set is 4 to 15 times longer than the longest runtime in each corresponding easy training set.  Validation RMSE on these hard problems was within 33\% of validation error when trained and validated on a random sampling for 3 of the 7 categories. This demonstrates that BPNNs have the potential to be trained on available data and then generalize to related problems that are too difficult for any current methods.  See the appendix for complete results.

\vspace{-0.1cm}
\paragraph{Learning Across Diverse Domains.}
We trained a BPNN on a random sampling of 70\% of problems from all categories, spanning network grid problems, bit-blasted versions of SMTLIB benchmarks, and network DQMR problems.  The BPNN achieved a final training RMSE of 3.9 and validation RMSE of 5.31, demonstrating that the BPNN is capable of capturing a broad distribution that spans multiple domains from a small training set.

\vspace{-0.1cm}
\section{Related Work}

\citep{abboud2020learning} use a graph neural network to perform approximate weighted disjunctive normal form (DNF) counting.  
Weighted DNF counting is a \#P-complete problem.  However, in contrast to model counting on CNF formulas, there exists an $O(nm)$ polynomial time approximation algorithm for weighted DNF counting (where $n$ is the number of variables and $m$ is the number of clauses).  The authors leverage this to generate a large training dataset of 100k DNF formulas with approximate solutions.  In comparison, our BPNN can learn and generalize from a very small training dataset of less than 50 problems.  This result provides the significant future work alluded to in the conclusion of \citep{abboud2020learning}.

Recently,\footnote{An early version of our paper concurrent with \citep{satorras2020neural} was submitted to UAI 2020: \\
\url{https://github.com/jkuck/jkuck.github.io/blob/master/files/BPNN_UAI_submission.pdf}} \citep{satorras2020neural} designed a graph neural network that operates on factor graphs and exchanges messages with BP to perform error correction decoding.  In contrast, BPNN-D preserves all of BP's fixed point, computes the exact partition function on tree structured factor graphs, and returns a lower bound whenever the Bethe approximation obtained from fixed points of BP is a provable lower bound.  All BPNN layers preserve BP's symmetries (invariances and equivariances) to permutations of both variable and factor indices.  Finally BPNN avoids `double counting' during message passing.

Prior work has shown that neural networks can learn how to solve NP-complete decision problems and optimization problems~\citep{selsam2018learning,prates2019learning,hsieh2019learning}.  \citep{Yoon2018InferenceIP} perform marginal inference in relatively small graphical models using GNNs.  \citep{heess2013learning} consider improving message passing in expectation propagation for probabilistic programming, when users can specify arbitrary code to define factors and the optimal updates are intractable. \citep{wiseman2019amortized} consider learning Markov random fields and address the problem of estimating marginal likelihoods (generally intractable to compute precisely).  They use a transformer network that is faster than LBP but computes comparable estimates.  This allows for faster amortized inference during training when likelihoods must be computed at every training step.  In contrast, BPNNs significantly outperform LBP and generalize to out of distribution data.%

\vspace{-0.1cm}
\section{Conclusion}

We introduced belief propagation neural networks, a strict generalization of BP that learns to find better fixed points faster.  The BPNN architecture resembles that of a standard GNN, but preserves BP's invariances and equivariances to permutations of variable and factor indices.  We empirically demonstrated that BPNNs can learn from tiny data sets containing only 10s of training points and generalize to test data drawn from a different distribution than seen during training.  BPNNs significantly outperform loopy belief propagation and standard graph neural networks in terms of accuracy.  BPNNs provide excellent computational efficiency, running orders of magnitudes faster than state-of-the-art randomized hashing algorithms while maintaining comparable accuracy.  

\section*{Broader impact}

This work makes both a theoretical contribution and a practical one by advancing the state-of-the-art in approximate inference on some benchmark problems. Our theoretical analysis of neural fixed point iterators is unlikely to have a direct impact on society. BPNN, on the other hand, can make approximate inference more scalable. Because approximate inference is a key computational problem underlying, for example, much of Bayesian statistics, it is applicable to many domains, both beneficial and harmful to society. Among the beneficial ones, we have applications of probabilistic inference to medical diagnosis and applications of model counting to reliability, safety, and privacy analysis.

\subsection*{Acknowledgements}
We thank Tri Dao, Ines Chami, and Shengjia Zhao for helpful discussions and feedback.  Research supported by NSF (\#1651565, \#1522054, \#1733686), ONR  (N00014-19-1-2145), AFOSR (FA9550-
19-1-0024), and FLI.

\bibliography{bpnn}
\bibliographystyle{plainnat}

\appendix

\section{PROOFS}
\label{appendix:proofs}

\begin{proof}[Theorem~\ref{thm:preserve_BP_fixed_points}]
Every fixed point of BP satisfies $\tilde{m}_{a \rightarrow i}^{(k)} = \overline{m}_{a \rightarrow i}^{(k-1)}$ by definition.  The computation of $\tilde{n}_{a \rightarrow i}^{(k)}$ from $\overline{n}_{a \rightarrow i}^{(k-1)}$ (Equations~\ref{eq:bpnn_varToFac_strong_guarantees} and \ref{eq:bpnn_facToVar_strong_guarantees}) is identical to the computation of $\tilde{m}_{a \rightarrow i}^{(k)}$ from $\overline{m}_{a \rightarrow i}^{(k-1)}$ in standard BP (Equations~\ref{eq:varToFacMsgs_log} and \ref{eq:facToVarMsgs_log}).  Therefore, every fixed point of BP satisfies $\overline{n}_{a \rightarrow i}^{(k-1)} = \tilde{n}_{a \rightarrow i}^{(k)}$ and is also a fixed point of BPNN-D when $H(0) = 0$
\end{proof}

\begin{proof}[Theorem~\ref{thm:no_new_fixed_points}]
Every fixed point of BPNN-D satisfies $\overline{n}_{a \rightarrow i}^{(k)} = \overline{n}_{a \rightarrow i}^{(k-1)}$ by definition.  Equation~\ref{eq:bpnn_varToFac_strong_guarantees} gives $\overline{n}_{a \rightarrow i}^{(k-1)} - \tilde{n}_{a \rightarrow i}^{(k)} =  \Delta^{(k)}_{a \rightarrow i} =  H\big(\overline{n}^{(k-1)} - \tilde{n}^{(k)}\big)_{a \rightarrow i}$.  %
Given the restriction on $H(\cdot)$ that $H(x)=x$ only if $x=0$, it follows that $\overline{n}_{a \rightarrow i}^{(k-1)} - \tilde{n}_{a \rightarrow i}^{(k)} = 0$.  This is a fixed point of BP by definition as the computation of $\tilde{n}_{a \rightarrow i}^{(k)}$ from $\overline{n}_{a \rightarrow i}^{(k-1)}$ is identical to the computation of $\tilde{m}_{a \rightarrow i}^{(k)}$ from $\overline{m}_{a \rightarrow i}^{(k-1)}$ in standard BP (Equations~\ref{eq:varToFacMsgs_log} and \ref{eq:facToVarMsgs_log}).
\end{proof}

\begin{proof}[Theorem~\ref{thm:BPNN-D_exact_LB}]
If zero is the unique fixed point of $H(\cdot)$, then the fixed points of BPNN-D and BP are identical by Theorems~\ref{thm:preserve_BP_fixed_points} and~\ref{thm:no_new_fixed_points}.  Therefore, (1) the Bethe approximation obtatined from fixed points of BPNN-D on tree structured factor graphs is exact because it is exact for fixed points of BP~\cite{koller2009probabilistic} (or see ~\cite{mori2013new}[p.27] for a detailed proof).  (2) \citet{Ruozzi2012TheBP}[p.8] prove in Corollary 4.2 that the Bethe approximation at any fixed point of BP is a lower bound on the partition function for factor graphs with binary variables and log-supermodular potential functions.  so it follows that the Bethe approximation at any fixed point of BPNN-D lower bounds the partition function.
\end{proof}

\begin{proof}[Proposition~\ref{prop:bpnn_convergence}]
If we consider a BPNN with weight tying, then regardless of the number of iterations or layers, the output messages are the same if the input messages are the same. Without loss of generality, let us first consider any node $r$ as the root node, and consider all the messages on the path from the leaf nodes through $r$. Let $d_{r, i}$ denote the depth of the sub-tree with root $i$ when we consider $r$ as the root (e.g. for a leaf node $i$, $d_{r, i} = 1$). We use the following induction argument:
\begin{itemize}
    \item At iteration $1$, the message from all nodes with $d_{r, i} = 1$ to their parents will be fixed for subsequent iterations since the inputs to the BPNN for these messages are the same.
    \item If at iteration $t - 1$, the message from all nodes with $d_{r, i} \leq t - 1$ to their parents are fixed for all subsequent iterations, then the inputs to the BPNN for all the messages from all nodes with $d_{r, i} = t$ to their parents will be fixed (since they depend on lower level messages that are fixed). Therefore, at iteration $t$, the messages from all the nodes with $d_{r, i} \leq t$ to their parents will be fixed because of weight tying between BPNN layers.
    \item The maximum tree depth is $l$, so $\max_i d_{r, i} \leq l$. From the induction argument above, after at most $l$ iterations, all the messages along the path from leaf nodes to $r$ will be fixed.
\end{itemize}
Since the BPNN layer performs the operation over all nodes, this above argument is valid for all nodes when we consider them as root nodes. Therefore, all messages will be fixed after at most $l$ iterations, which completes the proof.
\end{proof}

\paragraph{Isomorphic Factor Graphs}

To prove Theorem~\ref{thm:BPNN_symmetry_short} we define isomorphic factor graphs, an equivalence relation among factor graph representations, and break Theorem~\ref{thm:BPNN_symmetry_short} into the lemmas in this section.  Standard GNNs are built on the assumption that isomorphic graphs should be mapped to the same representation and non-isomorphic graphs should be mapped to different representations~\cite{xu2018powerful}.  This is a challenging goal, in fact~\cite{xu2018powerful}[p.4] prove in Lemma 2 that any GNN that aggregates messages from 1-hop neighbors is, at most, as discriminative as the Weisfeiler-Lehman (WL) graph isomorphism test.  \citet{xu2018powerful} go on to propose a provably `maximally powerful' GNN, one that maps isomorphic graphs to the same representation and maps non-isomorphic graphs to different representations whenever the WL test maps them to different representations, which is the best result possible for this class of graph neural networks that aggregate messages from 1-hop neighbors.  The input to a standard GNN is a graph represented by an adjacency matrix whose i-th row and column correspond to the i-th node.  Nodes and edges may have corresponding features.  The GNN in~\cite{xu2018powerful} was designed to map isomorphic graphs to the same representation by outputting learned node representations that are equivariant to the input node indexing and a graph wide representation that is invariant to input node indexing.

The input to a BPNN is a factor graph.  With the same motivation as for standard GNNs, BPNNs should map isomorphic factor graphs to the same output representation.  A factor graph is represented as\footnote{Note that a factor graph can be viewed as a weighted hypergraph where factors define hyperedges and factor potentials define hyperedge weights for every variable assignment within the factor.} $G = (A, F^p, F^{idx})$.  %
$A \in \{0,1\}^{M \times N}$ is an adjacency matrix over $M$ factor nodes and $N$ variable nodes, where$^{,}$\footnote{For readability, we use $a$ and $b$ to index factors and $i$ and $j$ to index variables throughout this section.} $A_{ai}=1$ if the i-th variable is in the scope of the a-th factor and $A_{ai}=0$ otherwise.  $F^p$ is an ordered list of $M$ factor potentials, where the a-th factor potential, $F_a^p$, corresponds to the a-th factor (row) in $A$ and is represented as a tensor with one dimension for every variable in the scope of $F^p_a$.  $F^{idx}$ is an ordered list of ordered lists that locally indexes variables within each factor.  $F^{idx}_a$ is an ordered list specifying the local indexing of variables within the a-th factor (in $A$ and $F^p$).  $F^{idx}_{ak} = i$ specifies that the k-th dimension of the tensor $F^p_a$ corresponds to the i-th variable (column) in $A$.  We define two factor graphs to be isomorphic when they meet the conditions of Definition~\ref{def:fg_isomorphism}.

\begin{definition}\label{def:fg_isomorphism}
Factor graphs $G = (G(A), G(F^p), G(F^{idx}))$ and $G' = (G'(A), G'(F^p), G'(F^{idx}))$ with $G(A) \in \{0,1\}^{M \times N}$ and $G'(A) \in \{0,1\}^{M' \times N'}$ are isomorphic if and only if %
$M=M'$, $N=N'$, and
\begin{enumerate}
    \item There exist bijections\footnote{For $K \in \mathbb{N}$, we use $[K]$ to denote $\{1, 2, \ldots, K\}$.}
    $f_F: [M] \to [M]$ and $f_V: [N] \to [N]$
    such that $G(A_{ai}) = G'(A_{bj})$ for all $a \in [M]$ and $i \in [N]$, where $b=f_F(a)$ and $j=f_V(i)$.
    \item There exists a bijection
    for every factor,
    \begin{equation}\label{eq:local_idx_bijection}
        f^{idx}_a : \{1, \dots, |G(F^{idx}_a)|\} \rightarrow \{1, \dots, |G'(F^{idx}_b)|\} \quad  \forall a \in [M],
    \end{equation}
    such that $f_V\big(G(F^{idx}_{ak})\big) = G'(F^{idx}_{bl})$  and $G(F^p_a) = \sigma_a\big(G'(F^p_b)\big)$, where where $b= f_F(a)$, $l=f^{idx}_a(k)$, $\sigma_a = \big((f^{idx}_a(1), f^{idx}_a(2), \dots, f^{idx}_a(|G(F^{idx}_a)|)\big)$, and $\sigma_a\big(G'(F^p_b)\big)$ denotes permuting the dimensions of the tensor $G'(F^p_b)$ according to $\sigma_a$.
\end{enumerate}
\end{definition}

Condition 1 in Definition~\ref{def:fg_isomorphism} states that permuting the \emph{global} indices of variables or factors in a factor graph results in an isomorphic factor graph.  Condition 2 in Definition~\ref{def:fg_isomorphism} states that permuting the \emph{local} indices of variables within factors also results in an isomorphic factor graph.  %
In Lemmas \ref{lemma:msg_equiv}, \ref{lemma:var_bel_equiv}, and \ref{lemma:fac_bel_equiv} we formalize the equivariance of messages and beliefs obtained by applying BPNN iterative layers.  We use using the bijections from Definition~\ref{def:fg_isomorphism} to construct bijective mappings %
between messages and beliefs.  In Lemma~\ref{lemma:BPNN-B_invariance} we use the equivariance of beliefs between isomorphic factor graphs to show that the output of BPNN-B is identical for isomorphic factor graphs.

\begin{lemma}\label{lemma:msg_equiv}
\textbf{Message equivariance:} Let $g^{(k)}_{i \rightarrow a}$ and $h^{(k)}_{i \rightarrow a}$ denote variable to factor messages and $g^{(k)}_{a \rightarrow i}$ and $h^{(k)}_{a \rightarrow i}$ factor to variable messages obtained by applying k iterations of BP to factor graphs $G$ and $G'$.  If $G$ and $G'$ are isomorphic as factor graphs and messages are initialized to a constant\footnote{Any message initialization strategy can be used, as long as initial messages are equivariant; e.g. they satisfy the bijective mapping  $g^{(0)}_{i \rightarrow a} = h^{(0)}_{j \rightarrow b}$ and $g^{(k)}_{a \rightarrow i} = h^{(k)}_{b \rightarrow j}$ where $j=f_V(i)$ and $b=f_F(a)$.} 
  
then there is a bijective mapping between messages: $g^{(k)}_{i \rightarrow a} = h^{(k)}_{j \rightarrow b}$ and $g^{(k)}_{a \rightarrow i} = h^{(k)}_{b \rightarrow j}$ where $j=f_V(i)$ and $b=f_F(a)$.  This property holds for BPNN-D iterative layers if $H(\cdot)$ is equivariant to global node indexing. 

\begin{proof}
We use a proof by induction.

Base case: the initial messages are all equal when constant initialization is used and therefore satisfy any bijective mapping.

Inductive step: 
Writing the definition of variable to factor messages, we have
\begin{equation}
    g_{i \rightarrow a}^{(k)}(x_i) = \prod_{c \in \mathcal{N}(i) \setminus a} g_{c \rightarrow i}^{(k-1)}(x_i) = \prod_{c \in \mathcal{N}(j) \setminus b} h_{c \rightarrow j}^{(k-1)}(x_j) = h^{(k)}_{j \rightarrow b}(x_j),
\end{equation}
since the bijective mapping holds for factor to variable messages at iteration $k-1$ by the inductive hypothesis.  Writing the definition of factor to variable messages, we have
\begin{align}\label{eq:FtoV_msg_equiv}
\begin{split}
    g_{a \rightarrow i}^{(k)}(x_i) = & \sum_{\mathbf{x}_a \setminus x_i} G(F^p_a)(\mathbf{x}_a) \prod_{l \in \mathcal{N}(a) \setminus i} g_{l \rightarrow a}^{(k)}(x_l) \\
    = & \sum_{\mathbf{x}_b \setminus x_j} \sigma_a\big(G'(F^p_b)\big)(\mathbf{x}_b) \prod_{l \in \mathcal{N}(b) \setminus j} g_{l \rightarrow b}^{(k)}(x_l) = g_{b \rightarrow j}^{(k)}(x_j). \\
\end{split}
\end{align}
showing that the bijective mapping continues to hold at iteration $k$.

Proof extension to BPNN-D: the logic of the proof is unchanged when BP is performed in log-space with damping.  The only difference between BPNN-D and standard BP is the replacement of the term $\alpha \big(\overline{m}_{a \rightarrow i}^{(k-1)} - \tilde{m}_{a \rightarrow i}^{(k)}\big)$ in the computation of factor to variable messages with $\Delta^{(k)}_{a \rightarrow i}$, where $\Delta^{(k)} = H\big(\overline{n}^{(k-1)} - \tilde{n}^{(k)}\big)$.  If $H(\cdot)$ is equivariant to global node indexing (the bijective mapping $\Delta^{(k)}_{a \rightarrow i}(G) = \Delta^{(k)}_{b \rightarrow j}(G')$ holds, where $\Delta^{(k)}_{a \rightarrow i}(G)$ denotes applying the operator $H(\cdot)$ to the k-th iteration's message differences when the input factor graph is $G$ and taking the output correpsonding to message $a \rightarrow i$), then equality is maintained in Equation~\ref{eq:FtoV_msg_equiv} and the bijective mapping between messages holds.
\end{proof}
\end{lemma}

\begin{lemma}\label{lemma:var_bel_equiv}
\textbf{Variable belief equivariance:} Let $g^{(k)}_i$ and $h^{(k)}_i$ denote the variable beliefs obtained by applying k iterations of BP (or BPNN-D iterative layers with $H(\cdot)$ equivariant to global node indexing) %
to factor graphs $G$ and $G'$.  If $G$ and $G'$ are isomorphic as factor graphs, then there is a bijective mapping between beliefs: $g^{(k)}_i = h^{(k)}_j$, where $j=f_V(i)$.
\begin{proof}
By the definition of variable beliefs,
\begin{equation}
    g_i^{(k)}(x_i) = \frac{1}{z_i} \prod_{a \in \mathcal{N}(i)} g_{a \rightarrow i}^{(k)}(x_i) = \frac{1}{z_j} \prod_{a \in \mathcal{N}(j)} h_{a \rightarrow j}^{(k)}(x_j) = h_j^{(k)}(x_j),
\end{equation}
where the second equality holds due to factor to variable message equivariance from Lemma~\ref{lemma:msg_equiv}.
\end{proof}
\end{lemma}

\begin{lemma}\label{lemma:fac_bel_equiv}
\textbf{Factor belief equivariance:} Let $g^{(k)}_a$ and $h^{(k)}_a$ denote the factor beliefs obtained by applying k iterations of BP
(or BPNN-D iterative layers with $H(\cdot)$ equivariant to global node indexing)
to factor graphs $G$ and $G'$.  If $G$ and $G'$ are isomorphic as factor graphs, then there is a bijective mapping between beliefs: $g^{(k)}_a = \sigma_a\big(h^{(k)}_b\big)$, where $b=f_F(a)$ and $\sigma_a = \big((f^{idx}_a(1), f^{idx}_a(2), \dots, f^{idx}_a(|G(F^{idx}_a)|)\big)$.
\begin{proof}
By the definition of factor beliefs,
\begin{align}
\begin{split}
    g_a^{(k)}(\mathbf{x}_a) =  & \frac{G(F^p_a)(\mathbf{x}_a)}{z_a} \prod_{i \in \mathcal{N}(a)} g_{i \rightarrow a}^{(k)}(x_i) = \frac{\sigma_a\big(G'(F^p_b)\big)(\mathbf{x}_b)}{z_b} \prod_{i \in \mathcal{N}(b)} h_{i \rightarrow b}^{(k)}(x_i) = \sigma_a\big(h^{(k)}_b(\mathbf{x}_b)\big), \\
\end{split}
\end{align}
where the second equality holds due to variable to factor message equivariance from Lemma~\ref{lemma:msg_equiv}.
\end{proof}
\end{lemma}

\begin{lemma}\label{lemma:BPNN-B_invariance}
\textbf{Bethe approximation invariance:} If factor graphs $G$ and $G'$ are isomorphic, then the Bethe approximations obtained by applying BP to $G$ and $G'$ (or the output of BPNN-B) are identical.
\begin{proof}
By the definition of the Bethe approximation (or the negative Bethe free energy),
\begin{align}
\begin{split}
    -F_{\textrm{Bethe}}(G) =  & \sum_{a=1}^M \sum_{\mathbf{x}_a} g_a(\mathbf{x}_a) \ln G(F^p_a)(\mathbf{x}_a) \\
    - & \sum_{a=1}^M \sum_{\mathbf{x}_a} g_a(\mathbf{x}_a) \ln   g_a(\mathbf{x}_a) + \sum_{i=1}^N (d_i - 1) \sum_{x_i} g_i(x_i) \ln g_i(x_i)\\
    = & \sum_{b=1}^M \sum_{\mathbf{x}_b} \sigma_{a'}(h_b)(\mathbf{x}_b) \ln \sigma_{a'}\big(G'(F^p_b)\big)(\mathbf{x}_b) \\
    - & \sum_{b=1}^M \sum_{\mathbf{x}_b} \sigma_{a'}(h_b)(\mathbf{x}_b) \ln   \sigma_{a'}(h_b)(\mathbf{x}_b) + \sum_{j=1}^N (d_j - 1) \sum_{x_j} h_j(x_j) \ln h_j(x_j) \\
    = & -F_{\textrm{Bethe}}(G')
\end{split}
\end{align}
where $a' = f_F^{-1}(b)$, the second equality follows from the equivariance of variable and factor beliefs (Lemmas~\ref{lemma:var_bel_equiv} and~\ref{lemma:fac_bel_equiv}), and the final equality follows from the commutative property of addition.

Proof extension to BPNN-B: the proof holds for BPNN-B because every permutation (in $S_{\lvert \mathbf{x}_a \rvert}$) of factor belief terms is input to $\text{MLP}_{BF}$.  
\end{proof}

\end{lemma}

\section{Extended Background}
We provide background on belief propagation and graph neural networks (GNN) to motivate and clarify belief propagation neural networks (BPNN).

\subsection{BELIEF PROPAGATION}

We describe a general version of belief propagation~\cite{yedidia2005constructing} that operates on factor graphs.

\paragraph{Factor Graphs.}
A factor graph~\cite{kschischang2001factor,yedidia2005constructing} is a general representation of a distribution over $n$ discrete random variables, $\{X_1, X_2, \dots, X_n\}$.  Let $x_i$ denote a possible state of the $i^{th}$ variable.  We use the shorthand $p(\mathbf{x}) = p(X_1 = x_1, \dots, X_n = x_1)$ for the joint probability mass function, where $\mathbf{x} = \{x_1, x_2, \dots, x_n\}$ is a specific realization of all $n$ variables.  Without loss of generality, $p(\mathbf{x})$ can be written as the product
\begin{equation} \label{eq:distribution_appendix}
    p(\mathbf{x}) = \frac{1}{Z}  \prod_{a=1}^M f_a(\mathbf{x}_a).
\end{equation}
The functions $f_1, f_2, \dots, f_m$ each take some subset of variables as arguments; function $f_a$ takes $\mathbf{x_a} \subset \{x_1, x_2, \dots, x_n\}$.  We require that all functions are non-negative and finite.  This makes $p(\mathbf{x})$ a well defined probability distribution after normalizing by the distribution's partition function
\begin{equation}
    Z = \sum_{\mathbf{x}} \left ( \prod_{a=1}^M f_a(\mathbf{x}_a) \right ).
\end{equation}
A factor graph is a bipartite graph that expresses the factorization of the distribution in equation~\ref{eq:distribution_appendix}.  A factor graph's nodes represent the $n$ variables and $M$ functions present in equation~\ref{eq:distribution_appendix}.  The nodes corresponding to functions are referred to as factor nodes.  Edges exist between factor nodes and variables nodes if and only if the variable is an argument to the corresponding function.

\paragraph{Message Updates.}

Belief propagation performs iterative message passing.  The message $m_{i \rightarrow a}^{(k)}(x_i)$ from variable node $i$ to factor node $a$ during iteration $k$ is computed according to the rule
\begin{equation} \label{eq:varToFacMsgs}
    m_{i \rightarrow a}^{(k)}(x_i) \coloneqq \prod_{c \in \mathcal{N}(i) \setminus a} m_{c \rightarrow i}^{(k-1)}(x_i).
\end{equation}
The message $m_{a \rightarrow i}^{(k)}(x_i)$ from factor node $a$ to variable node $i$ during iteration $k$ is then computed according to the rule
\begin{equation} \label{eq:facToVarMsgs}
    m_{a \rightarrow i}^{(k)}(x_i) \coloneqq \sum_{\mathbf{x}_a \setminus x_i} f_a(\mathbf{x}_a) \prod_{j \in \mathcal{N}(a) \setminus i} m_{j \rightarrow a}^{(k)}(x_j).
\end{equation}

The BP algorithm estimates approximate marginal probabilities for each variable, referred to as beliefs.  We denote the belief at variable node $i$, after message passing iteration $k$ is complete, as $b_i^{(k)}(x_i)$ which is computed as 
\begin{equation}
    b_i^{(k)}(x_i) = \frac{1}{z_i} \prod_{a \in \mathcal{N}(i)} m_{a \rightarrow i}^{(k)}(x_i), \text{with normalization } z_i = \sum_{x_i} \prod_{a \in \mathcal{N}(i)} m_{a \rightarrow i}^{(k)}(x_i).
\end{equation}
Similarly, BP computes joint beliefs over the sets of variables $\mathbf{x}_a$ associated with each factor $f_a$.  We denote the belief over variables $\mathbf{x}_a$, after message passing iteration $k$ is complete,  as $b_a^{(k)}(\mathbf{x}_a)$ which is computed as
\begin{equation}
    b_a^{(k)}(\mathbf{x}_a) =  \frac{f_a(\mathbf{x}_a)}{z_a} \prod_{i \in \mathcal{N}(a)} m_{i \rightarrow a}^{(k)}(x_i), \text{with normalization } z_a = \sum_{\mathbf{x}_a} f_a(\mathbf{x}_a) \prod_{i \in \mathcal{N}(a)} m_{i \rightarrow a}^{(k)}(x_i).
\end{equation}

\paragraph{Partition Function Approximation.}
The belief propagation algorithm proceeds by iteratively updating variable to factor messages (Equation~\ref{eq:varToFacMsgs}) and factor to variable messages (Equation~\ref{eq:facToVarMsgs}) until they converge to fixed values, referred to as a fixed point of Equations~\ref{eq:varToFacMsgs} and \ref{eq:facToVarMsgs}, or a predefined maximum number of iterations is reached.  While BP is not guaranteed to converge in general, whenever a fixed point is found it defines a set of consistent beliefs, meaning that marginal beliefs at factor nodes agree with beliefs every variable node they are connected to.  At this point the beliefs are used to compute a variational approximation of the factor graph's partition function.  This approximation, originally developed in statistical physics, is known as the Bethe free energy $F_{\textrm{Bethe}}\approx -\ln Z$~\cite{bethe1935statistical}.  It is defined in terms of the Bethe average energy $U_{\textrm{Bethe}}$ and the Bethe entropy $H_{\textrm{Bethe}}$.

\begin{definition}
$U_{\textrm{Bethe}} \coloneqq - \sum_{a=1}^M \sum_{\mathbf{x}_a} b_a(\mathbf{x}_a) \ln f_a(\mathbf{x}_a)$ defines the Bethe average energy.
\end{definition}

\begin{definition}
 $H_{\textrm{Bethe}} \coloneqq  - \sum_{a=1}^M \sum_{\mathbf{x}_a} b_a(\mathbf{x}_a) \ln   b_a(\mathbf{x}_a) + \sum_{i=1}^N (d_i - 1) \sum_{x_i} b_i(x_i) \ln b_i(x_i)$ defines the Bethe entropy,
where $d_i$ is the degree of variable node $i$.
\end{definition}

\begin{definition}\label{eq:bethe_free_energy}
The Bethe free energy is defined as ${F_{\textrm{Bethe}} = U_{\textrm{Bethe}} - H_{\textrm{Bethe}}}$.
\end{definition}

\subsection{GNN Background}
\label{Appendix_GNN}
This section provides background on %
graph neural networks (GNNs), a form of neural network used to perform representation learning on graph structured data.  GNNs perform iterative message passing operations between neighboring nodes in graphs, updating the learned, hidden representation of each node after every iteration.  %
\citet{xu2018powerful} showed that graph neural networks are at most as powerful as the Weisfeiler-Lehman graph isomorphism test~\cite{weisfeiler1968reduction}, which is a strong test that generally works well for discriminating between graphs.  Additionally, \citep{xu2018powerful} presented a GNN architecture called the Graph Isomorphism Network (GIN), which they showed has discriminative power equal to that of the Weisfeiler-Lehman test and thus strong representational power.  We will use GIN as a baseline GNN for comparison in our experiments because it is provably as discriminative as any GNN that aggregates information from 1-hop neighbors.

We now describe in detail the GIN architecture that we use as a baseline. Our architecture performs regression on graphs, learning a function $f_{\text{GIN}}: \mathcal{G} \rightarrow \mathbb{R}$ from graphs to a real number.  Our input is a graph $G=(V, E) \in  \mathcal{G}$ with node feature vectors $\mathbf{h}_v^{(0)}$  for $v \in V$ and edge feature vectors $\mathbf{e}_{u,v}$ for $(u,v) \in E$.  Our output is the number $f_{\text{GIN}}(G)$, which should ideally be close to the ground truth value $y_G$. Let $\mathbf{h}_v^{(k)}$ denote the representation vector corresponding to node $v$ after the $k^{th}$ message passing operation.  %
We use a slightly modified GIN update to account for edge features as follows: %
\begin{align}\label{eq:gin_edge_update}
\begin{split}
    \mathbf{h}_v^{(k)} = \text{MLP}_1^{(k)} 
        \Bigg(  \mathbf{h}_v^{(k-1)} + 
        \sum_{u \in \mathcal{N}(v)} \text{MLP}_2^{(k)} \left(\mathbf{h}_u^{(k-1)}, \mathbf{e}_{u,v} \right) \Bigg).
\end{split}
\end{align}

A $K$-layer GIN network with width $M$ is defined by $K$ successive GIN updates as given by Equation~\ref{eq:gin_edge_update}, where  $\mathbf{h}_v^{(k)} \in \mathbb{R}^M$ is an $M$-dimensional feature vector for $k \in \{1,2,\dots,K\}$.  All MLPs within GIN updates (except $\text{MLP}_2^{(0)}$) are multilayer perceptrons with a single hidden layer whose input, hidden, and output layers all have dimensionality $M$.  $\text{MLP}_2^{(0)}$ is different in that its input dimensionality is given by the dimensionality of the original node feature representations.  The final output of our GIN network is given by
\begin{equation}\label{eq:gin_final_layer}
    f_{\text{GIN}}(G) =  \text{MLP}^{(K+1)} \left( \CONCAT{k=1}{K} \sum_{v \in G} \mathbf{h}_v^{k} \right),
\end{equation}
where we concatenate summed node feature vectors from all layers and $\text{MLP}^{(K+1)}$ is a multilayer perceptron with a single hidden layer.  Its input and hidden layers have dimensionality $M \cdot K$ and its output layer has dimensionality 1.

\section{BPNN Iterative Layer Additional Variants}
\label{Appendix_BPNN}

When the convergence properties of BPNN-D are not needed (e.g., if BP is not a lower bound to the partition function of a particular problem), we have the flexibility to create BPNN iterative layers that directly operate on a combination of messages and beliefs by modifying $\tilde{m}_{i \rightarrow a}^{(k)}$ and $\tilde{m}_{a \rightarrow i}^{(k)}$ from Equations \ref{eq:varToFacMsgs_log} and \ref{eq:facToVarMsgs_log}. We can introduce a variant that parameterizes both factor to variable messages and factor beliefs and computes factor to variable messages as:

\begin{equation} \label{bpnn_mlp1_2}
\tilde{m}_{a \rightarrow i}^{(k)}= -z_{a \rightarrow i} + \LSE{\mathbf{x}_a \setminus x_i} \Bigg(  \phi_a(\mathbf{x}_a) + \LNE_2 \bigg[ \sum_{j \in \mathcal{N}(a) \setminus i} \LNE_1 \Big( \overline{m}_{j \rightarrow a}^{(k)}) \bigg] \Bigg)
\end{equation} 

where we use the shorthand

\begin{equation} \label{eq:logexp_mlp_residual}
    \LNE_{i} (\mathbf{h}) = \ln \Big( \text{MLP}_{\theta_{i}} \big( \exp (\mathbf{h}) \big) \Big),
\end{equation}
and $\text{MLP}_{\theta_{i}}$ is a multilayer perceptron parameterized by $\theta_{i}$. We exponentiate before applying the multilayer perceptron because we empirically find that this improves training as opposed to having MLPs operate directly in log space. 

We can also introduce additional variants that operate only on messages and parameterize both variable to factor and factor to variable messages:

\begin{equation} \label{eq:varToFacMsgs_bpnn}
     \tilde{m}_{i \rightarrow a}^{(k)} = -z_{i \rightarrow a} + \sum_{c \in \mathcal{N}(i) \setminus a} LNE_{3}(\overline{m}_{c \rightarrow i}^{(k-1)}).
\end{equation}

\begin{equation} \label{eq:facToVarMsgs_bpnn}
    \tilde{m}_{a \rightarrow i}^{(k)}= -z_{a \rightarrow i} + \LSE{\mathbf{x}_a \setminus x_i} \bigg( \phi_a(\mathbf{x}_a) + \sum_{j \in \mathcal{N}(a) \setminus i} LNE_{4}(\overline{m}_{j \rightarrow a}^{(k)}) \bigg),
\end{equation}

BPNN iterative layers allow for great flexibility and different combinations of these MLPs can be applied in a specific layer depending on the task at hand. These MLPs can even be combined with the damping MLPs found in BPNN-D layers in lieu of the fixed scalar damping coefficient $\alpha$ found in Equations \ref{eq:varToFacMsgs_log} and \ref{eq:facToVarMsgs_log}.

\paragraph{BPNN Initialization}
Note that any BPNN architecture built from iterative layers with or without a BPNN-B layer can be initialized to perform BP run for a fixed number of iterations by initializing MLPs functions $f(x) = x$.  E.g. weight matrices are set to the identity, bias terms to zero, and any nonlinearities are chosen so as to avoid affecting the input at initialization.

\section{Ising Model Experiments}

\paragraph{Data Generation.}
An $N \times N$ Ising model is defined over binary variables $x_i \in \{-1,1\}$ for $i=1,2,\dots,N^2$, where each variable represents a spin.  Each spin has a local field parameter $J_i$ which corresponds to its local potential function $J_i(x_i) = J_i x_i$.  Each spin variable has 4 neighbors, unless it occupies a grid edge.  Neighboring spins interact with coupling potentials $J_{i,j}(x_i, x_j) = J_{i,j} x_i x_j$.  The probability of a complete variable configuration $\mathbf{x} = \{x_1, \dots, x_{N^2}\}$ is defined to be 
\begin{equation}
    p(\mathbf{x}) = \frac{1}{Z} \exp \left( \sum_{i \in V} J_i x_i + \sum_{(i,j) \in E} J_{i,j} x_i x_j \right),
\end{equation}
where the normalization constant $Z$, or partition function, is defined to be
\begin{equation}
    Z = \sum_{\mathbf{x}} \exp \left( \sum_{i \in V} J_i x_i + \sum_{(i,j) \in E} J_{i,j} x_i x_j \right).
\end{equation}

We performed experiments using datasets of randomly generated Ising models.  Each dataset was created by first choosing $N$, $c_\text{max}$, and $f_\text{max}$.  %
We sampled $N \times N$ Ising models according to the following process
\begin{align*}
& c \sim \mathrm{Unif}[0, c_\text{max}), \\
& f \sim \mathrm{Unif}[0, f_\text{max}), \\
& (J_i)_{i \in V} \iidsim \mathrm{Unif}[-f, f), \\
& (J_{i,j})_{(i,j) \in E} \iidsim \mathrm{Unif}[0, c).\\
\label{eq:ising_model_attr}
\end{align*}

\paragraph{Baselines.}
We trained a 10 layer GNN (GIN architecture) with width 4 on the same dataset of attractive Ising models that we used for our BPNN.  We set edge features to the coupling potentials; that is, $\mathbf{e}_{u,v} = J_{u,v}$.  We set the initial node representations to the local field potentials of each node, $\mathbf{h}_v^{(0)} = J_v$.  We used the same training loss and optimizer as for our BPNN.  We used an initial learning rate of 0.001 and trained for 5k epochs, decaying the learning rate by .5 every 2k epochs.

We consider two additional baselines: Bethe approximation from running standard loopy belief propagation and mean field approximation.  We used the libDAI~\cite{Mooij_libDAI_10} implementation for both.  We test loopy belief propagation with and without damping and with both parallel and sequential message update strategies.  We show results for two settings whose estimates of the partition function differ most drastically: (1) run for a maximum of 10 iterations with parallel updates and damping set to .5, and (2) run for a maximum of 1000 iterations with sequential updates using a random sequence and no damping.

\begin{figure}[htp]
\centering
\includegraphics[width=.3\textwidth]{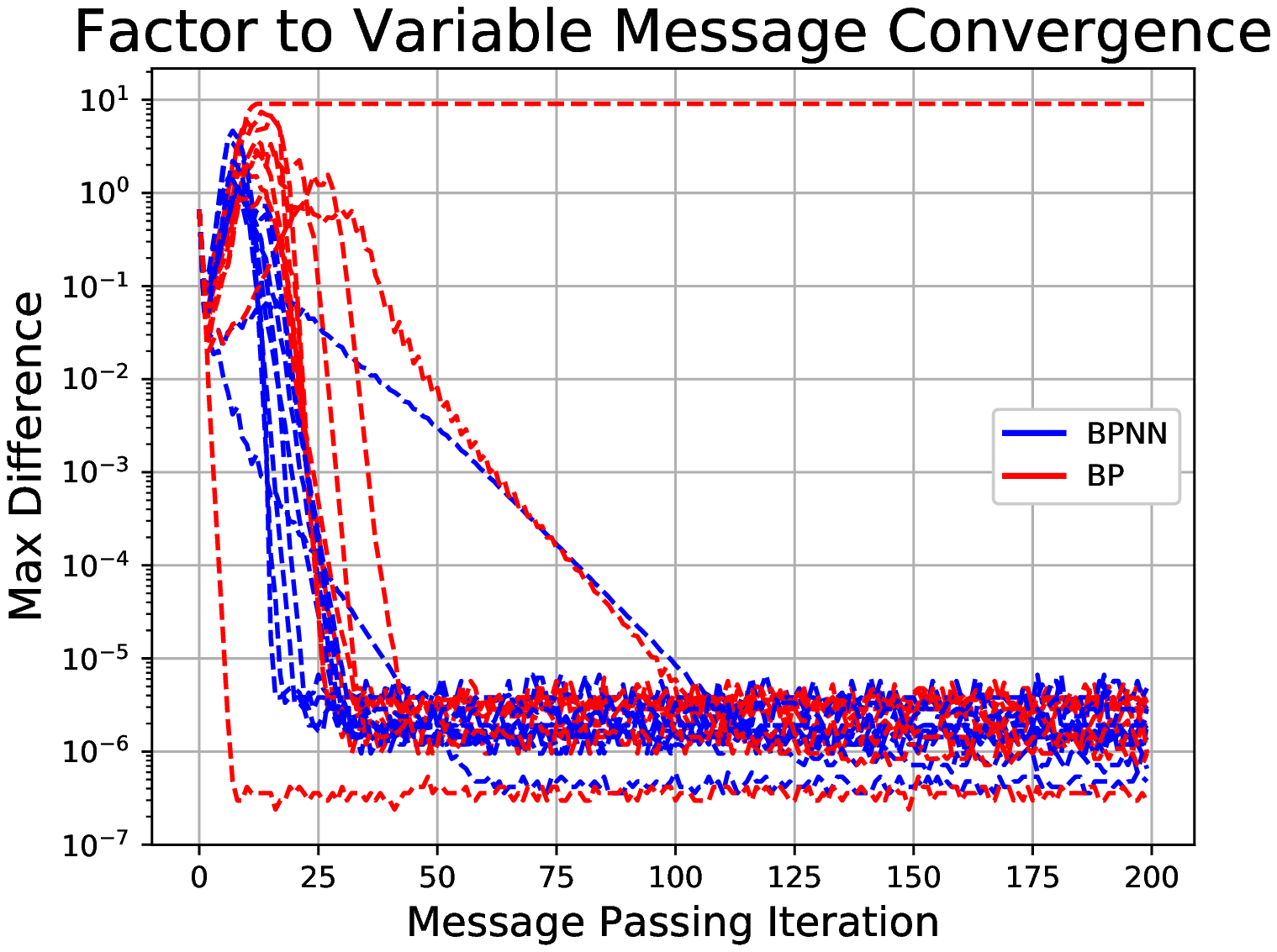}\quad
\includegraphics[width=.3\textwidth]{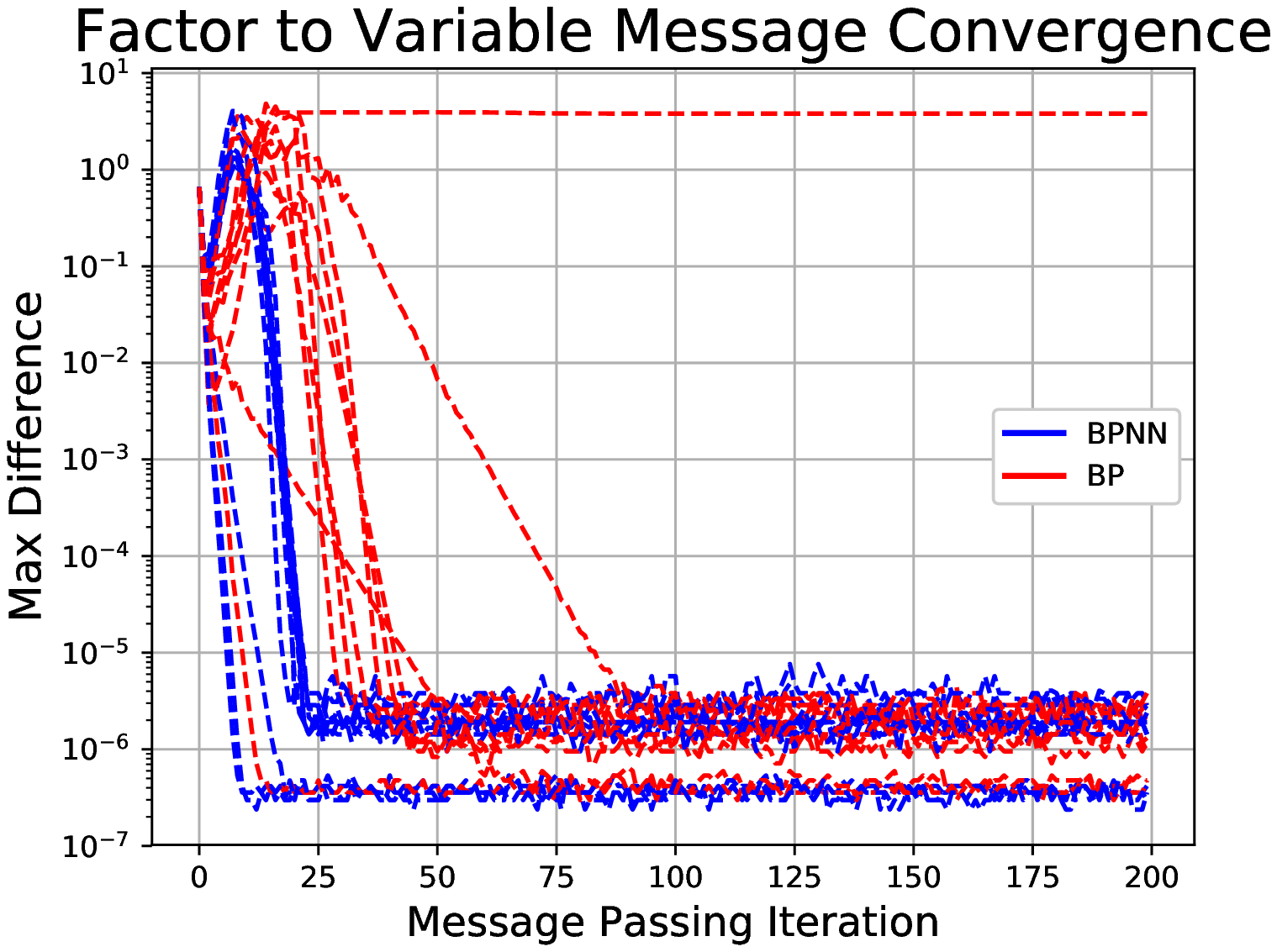}\quad
\includegraphics[width=.3\textwidth]{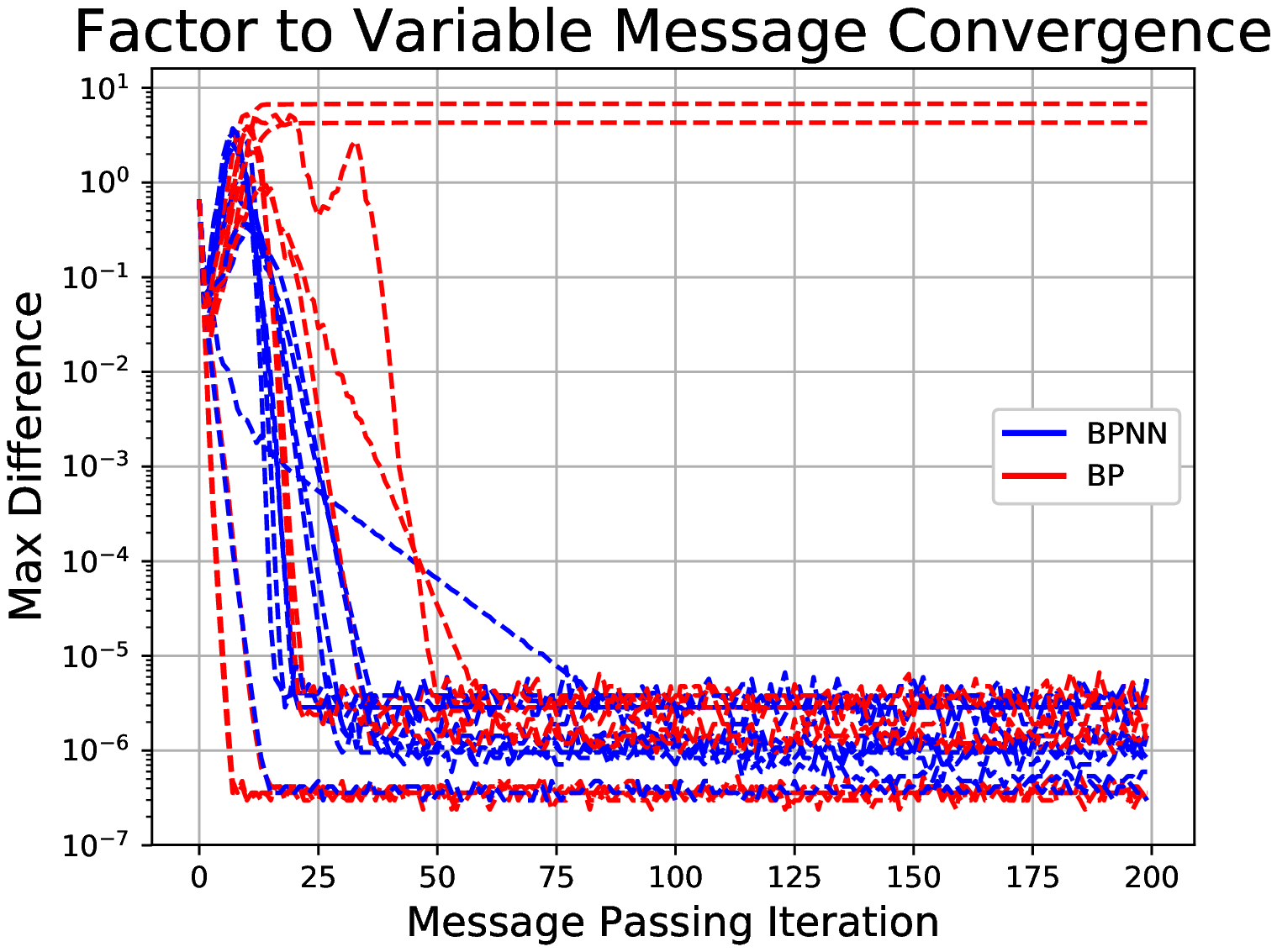}

\medskip

\includegraphics[width=.3\textwidth]{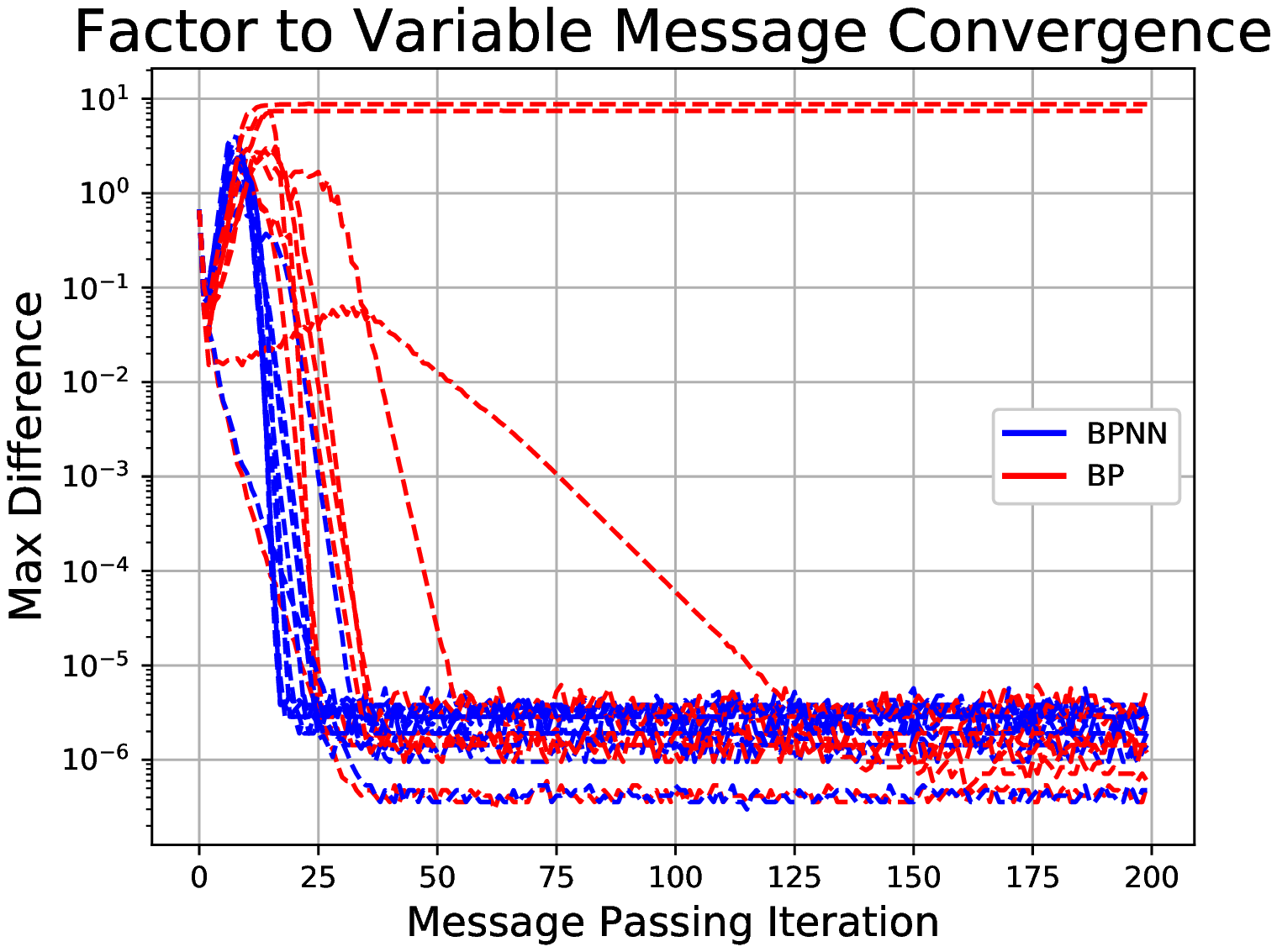}\quad
\includegraphics[width=.3\textwidth]{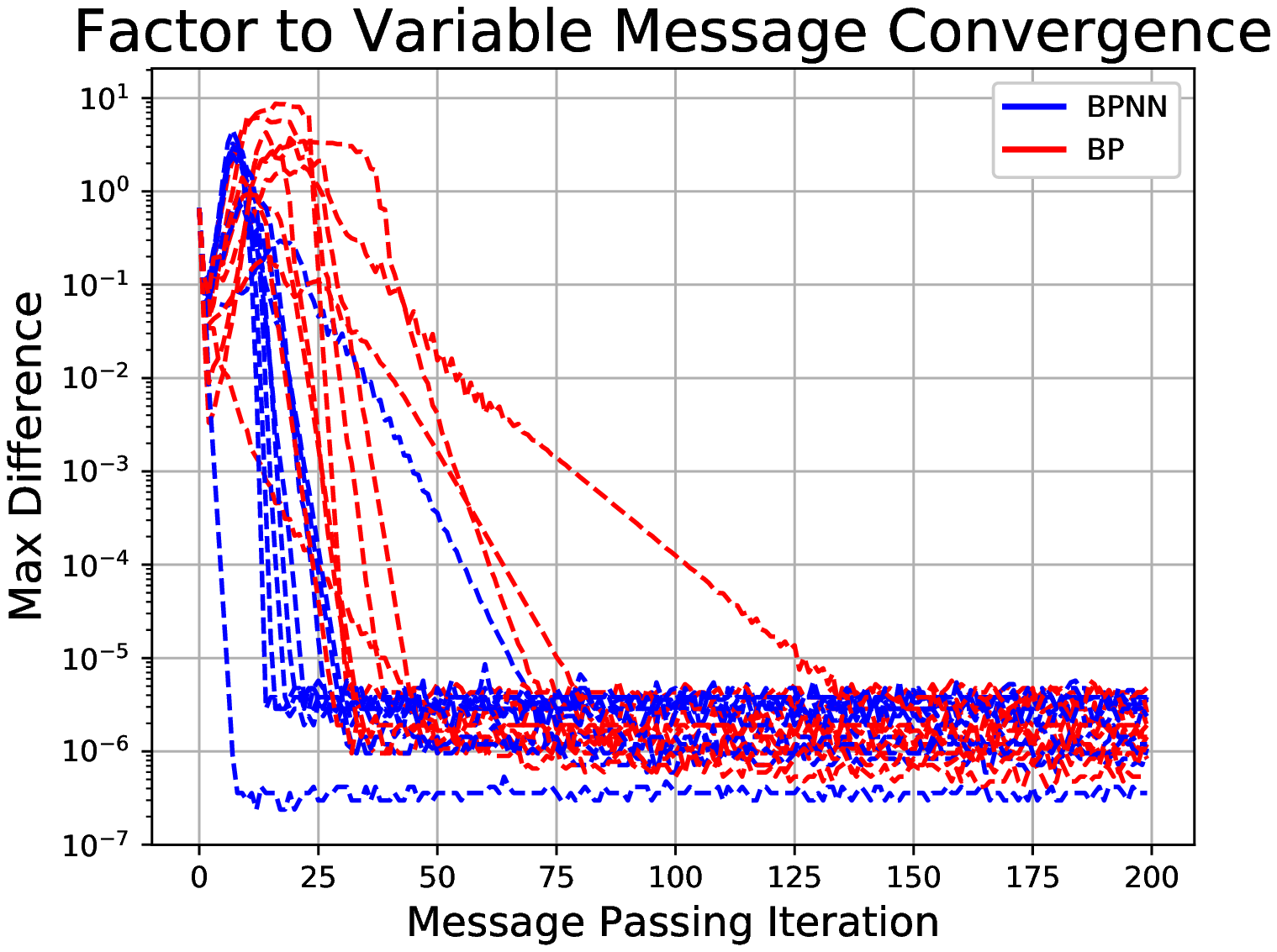}

\caption{The maximum difference in factor to variable message values between iterations is plotted against the message passing iteration for BPNN-D and BPNN run on 50 validation Ising models. BPNN-D converges to a maximum difference of $10^{-5}$ faster than BP, with a median speedup of 1.7x.}
\label{fig:BPNN-D_convergence}
\end{figure}

\paragraph{Improved Lower Bounds and Faster Convergence.}
We trained a BPNN-D to estimate the partition function on a training set of 50 random Ising models.  We randomly sampled the number of iterations of BPNN-D to apply during training between 5 and 30.  When BPNN-D is then run to convergence on a validation set of random Ising models, we find that (1) it finds fixed points that provide tighter lower bounds on the partition function as explained in the main text and (2) it converges faster than BP as shown in Figure~\ref{fig:BPNN-D_convergence}. 

\paragraph{Out of Distribution Generalization.}
We tested BPNN's ability to generalize to larger factor graphs and to shifts in the test distribution.  Again we used a training set of 50 Ising models.
We sampled test data from distributions with $c_{max}$ and $f_{max}$ increased by factors of 2 and 10 from their training values, with $N$ set to 14 (for 196 variables instead of the 100 seen during training). For this experiment we used a BPNN architecture with 10 iterative layers whose weights were not tied and with MLPs that operate on factor messages.  
For the out of distribution experiments, we did not use a final BPNN-B layer, we set the residual parameters to $\alpha_0=\alpha_1=\alpha_2=.5$, and trained on 50 attractive Ising models generated with $N=10$, $f_\text{max}=.1$, and $c_\text{max}=5$.  We used mean squared error as our training loss.  We used the Adam optimizer~\cite{Kingma2015adam} with an initial learning rate of .0005 and trained for 100 epochs, with a decay of .5 after 50 epochs.  Batching was over the entire training set (of size 50) with one optimization step per epoch.

\section{SAT Experiments}
\paragraph{Additional Dataset Details}
We evaluated the performance of our BPNN using the suite of benchmarks from \citet{SM19}.  Some of these benchmarks come with a sampling set.  The sampling set redefines the model counting problem, asking how many configurations of variables in the sampling set correspond to at least one complete variable configuration that satisfies the formula.  (A formula with $n$ variables may have at most $2^n$ satisfying solutions, but a sampling set over $i$ variables will restrict the number of solutions to at most $2^i$).  We stripped all problems of sampling sets since they are outside the scope of this work.  We also stripped all problems of minimal independent support variables sets and recomputed these when possible (we will discuss further later in this section).We ran the exact model counter DSharp\footnote{\url{https://github.com/QuMuLab/dsharp}}~\cite{Muise2012} on all benchmarks with a timeout of 5k seconds to obtain ground truth model counts for 928 of the 1,896 benchmarks.  Only 50 of these problems had more than 5 variables in the largest factor, so we discarded these problems and set the BPNN architecture to run on factors over 5 variables.  We categorized the remaining 878 by their arcane names into groupings.  With some sleuthing we determined that categories `or\_50', `or\_60', `or\_70', and `or\_100' contain network DQMR problems  with 150, 121, 111, and 138 benchmarks per category respectively.  Categories `75' and `90' contain network grid problems with 20 and 107 benchmarks per category respectively.  Category `blasted' conains bit-blasted versions of SMTLIB ( satisfiability modulo theories library) benchmarks~\cite{CMV16} and has 147 benchmarks.  Category `s' contains representations of circuits with a subset of outputs randomly xor-ed and has 68 benchmarks.  %
We discarded 4 categories that contained fewer than 10 benchmarks.  For each category that contained more than 10 benchmarks, we split 70\% into the training set and left the remaining benchmarks in the test set.  We then performed two splits of the training set for training and validation; for each category we (1) trained on a random sampling of 70\% of the training problems and performed validation on the remaining 30\% and (2) trained on 70\% of the training problems that DSharp solved fastest and performed validation on the remaining 30\% that took longest for DSharp to solve.  These hard validation sets are significantly more challenging for Dsharp.  The median runtime in each category's hard validation set is 4 to 15 times longer than the longest runtime in each corresponding easy training set.

\paragraph{Minimal Independent Support}
As a pre-processing step for ApproxMC3 and F2, we attempted to find a set of variables that define a \textit{minimal independent support} (MIS)~\cite{ivrii2016computing} for each benchmark using the authors' code\footnote{\url{https://github.com/meelgroup/mis}} with a timeout of 1k seconds.  A set of variables that define a MIS for a boolean formula fully determine the values of the remaining variables.  Randomized hashing algorithms can run significantly faster when given a set of variables that define a MIS.  When we could find a set of variables that define a MIS, we recorded the time that each randomized hashing algorithm required without the MIS and the sum of the time to find the MIS and perform randomized hashing with the MIS.  We report the minimum of these two times.  

\paragraph{Baseline Approximate Model Counters.}
For comparison, we ran the state of the art approximate model counter ApproxMC3\footnote{\url{https://github.com/meelgroup/ApproxMC}}~\citep{CMV16,SM19} on all benchmarks.  ApproxMC3 is a randomized hashing algorithm that returns an estimate of the model count that is guaranteed to be within a multiplicative factor of the exact model count with high probability.  Improving the guarantee, either by tightening the multiplicative factor or increasing the confidence, will increase the algorithm's runtime.  We ran ApproxMC3 with the default parameters; confidence set to 0.81 and epsilon set to 16.  

We also compare with the state of the art randomized hashing algorithm F2\footnote{\url{https://github.com/ptheod/F2}} from \citep{achlioptas2017probabilistic,achlioptas2018fast}, run with CryptoMiniSat5\footnote{\url{https://github.com/msoos/cryptominisat}}~\cite{Soos2009ExtendingSS,SM19}.  This algorithm gives up the probabilistic guarantee that the returned estimate will be within a multiplicative factor of the true model count in return for significantly increased computational efficiency.  We computed only a lower bound and ran F2 with variables appearing in only 3 clauses.  This significantly speeds up the reported results~\citep[p.14]{achlioptas2017probabilistic}, at some additional cost to accuracy.  For example, on the problem `blasted\_case37' ~\cite[p.14]{achlioptas2017probabilistic} report an estimate of $\log_2(\# \textrm{models}) \approx 151.02$ and a runtime of 4149.9 seconds.  Running F2 with variables appearing in only 3 clauses, we computed the lower bound on $\log_2(\# \textrm{models})$ of 148 in 2 seconds.

We also attempted to train a GNN, using the architecture from~\citep{selsam2018learning} to perform regression instead of classification.  We used the author's code, making slight modifications to perform regression.  However, we were not successful in achieving non-trivial learning.

\begin{table}[t!]
\centering
\begin{tabular}{@{}l c c c c @{}}
\toprule
\multicolumn{5}{c}{\textbf{RMSE Ablation Study by SAT Category}}  \\
Benchmark &  Train / Val  & Train / Val & Train / Val  & Train / Val \\
Category &   Split &   BPNN &    BPNN-NI  &   BPNN-DC \\
\midrule
\multirow{ 2}{*}{`or\_50'} & Random Split &.32 / \textbf{.30} &0.18 / 0.37 &0.95 / 2.21 \\
& Easy / Hard &.31 / 1.10 &0.17 / \textbf{0.70} &0.72 / 4.81 \\
\midrule
\multirow{ 2}{*}{`or\_60'} & Random Split &.32 / \textbf{.39} &0.20 / 0.43 &0.70 / 2.36 \\
& Easy / Hard &.31 / 1.7 &0.21 / \textbf{1.22} &0.57 / 3.69 \\
\midrule
\multirow{ 2}{*}{`or\_70'} & Random Split &.28 / .53 &0.19 / \textbf{0.45} &0.79 / 1.85 \\
& Easy / Hard &.35 / \textbf{.50} &0.19 / 0.58 &0.73 / 2.75 \\
\midrule
\multirow{ 2}{*}{`or\_100'} & Random Split &.48 / \textbf{.51} &0.31 / 0.59 &1.05 / 2.37 \\
& Easy / Hard &.48 / \textbf{.58} &0.27 / 0.60 &0.89 / 288.91 \\
\midrule
\multirow{ 2}{*}{`blasted'} & Random Split &4.39 / \textbf{4.24} &4.22 / 10.39 &3.01 / 6.59 \\
& Easy / Hard &2.19 / 10.10 &1.51 / \textbf{8.28} &1.57 / 948.95 \\
\midrule
\multirow{ 2}{*}{`75'} & Random Split &1.69 / 1.39 &1.31 / 1.34 &0.66 / \textbf{0.36} \\
& Easy / Hard &1.51 / 2.81 &0.98 / 2.96 &0.65 / \textbf{2.00} \\
\midrule
\multirow{ 2}{*}{`90'} & Random Split &2.46 / 2.18 &1.59 / 2.21 &0.94 / \textbf{1.59} \\
& Easy / Hard &2.17 / 2.86 &1.76 / 3.27 &0.81 / \textbf{1.23} \\
\midrule
All Categories & Random Split & 3.92 / \textbf{5.31} & 6.77 / 9.57 & 4.28 / 27.03\\
\bottomrule
\end{tabular}
\caption{RMSE of BPNN for each training/validation set, along with ablation results.  BPNN corresponds to a model with 5 BPNN-D layers followed by a Bethe layer that is invariant to the factor graph representation.  BPNN-NI corresponds to removing invariance from the Bethe layer.  BPNN-DC corresponds to performing 'double counting' as is standard for GNN, rather than subtracting previously sent messages as is standard for BP.  'Random Split' rows show that BPNNs are capable of learning a distribution from a tiny dataset of only 10s of training problems.  `Easy / Hard' rows additionally show that BPNNs are able to generalize from simple training problems to significantly more complex validation problems.}\label{tab:bpnn_sat}

\end{table}

\paragraph{BPNN Training Protocol.}

We trained our BPNN to predict the natural logarithm of the number of satisfying solutions to an input boolean formula.  We consider the general case of an input formula over $n$ boolean variables, $\{X_1, X_2, \dots, X_n\}$, in conjunctive normal form (CNF).  Formulas in CNF are a conjunction of clauses, where each clause is a disjunction of literals.  A literal is either a variable or its negation.  We converted boolean formulas into factor graphs where each clause corresponds to a factor.  Factors take the value of 1 for variable configurations that satisfy the clause and 0 for variable configurations that do not satisfy the clause.  The partition function of this factor graph is equal to the number of satisfying solutions.  %
We trained a BPNN architecture composed of 5 BPNN-D layers followed by a BPNN-B layer.  We used the Adam optimizer~\cite{Kingma2015adam} with learning rate decay.

\paragraph{Ablation Study.}
The columns labeled BPNN-NI and BPNN-DC in table~\ref{tab:bpnn_sat} correspond to ablated versions of our BPNN model.  We trained a BPNN with a BPNN-B layer that was not permutation invariant to local variable indexing, by removing the sum over permutations in $S_{\lvert \mathbf{x}_a \rvert}$ from Equation~\ref{eq:bethe_MLP} and only passing in the original beliefs.  We refer to this non-invariant version as BPNN-NI. We then forced BPNN-NI to `double count' messages by changing the sums in Equations~\ref{eq:bpnn_varToFac_strong_guarantees} and \ref{eq:bpnn_facToVar_strong_guarantees} to be over $j \in \mathcal{N}(a)$.  We this non-invariant version that performs double counting as BPNN-DC.  We observe validation improvement in BPNN over these ablated versions when generalization is particularly challenging, e.g. on `blasted' problems individually and on all categories.

\paragraph{Additional Baseline Approximate Model Counter Information}

Table~\ref{tab:baselines_sat} shows the root mean squared error (RMSE) of estimates from the approximate model counters ApproxMC3 and F2 across all training benchmarks in each category.  Error was computed as the difference between the natural logarithm of the number of satisfying solutions and the estimate.  The fraction of benchmarks that each approximate counter was able to complete within the time limit of 5k seconds is also shown.  For each benchmark category we show runtime percentiles for ApproxMC3, F2, and the exact model counter DSharp in Table~\ref{tab:runtimes_sat}.  The DSharp runtime column shows the runtime dividing our easy training sets and hard validation sets for each benchmark category.  It also shows the median run time of each hard validation set (85th percentile).  The median runtime in each category's hard validation set is 4 to 15 times longer than the longest runtime in each corresponding easy training set.  We observe that F2 is generally tens or hundreds of times faster than ApproxMC3.  On these benchmarks DSharp is generally faster than F2, however there exist problems that can be solved much faster by randomized hashing (ApproxMC3 or F2) than by DSharp~\cite{achlioptas2018fast,SM19}.

\begin{table}[t]
\centering
\begin{tabular}{@{}l cc @{}}
\toprule
\multicolumn{3}{c}{\textbf{Baselines RMSE by SAT Category}}  \\
  & \multicolumn{2}{c}{RMSE (\% Completed)} \\
 \cmidrule(lr){2-3} 
  Category &  ApproxMC3  & F2  \\
\midrule
`or\_50' & 0.07 (89\%) & 2.4 (100\%) \\
`or\_60' & 0.07 (87\%) & 2.3 (100\%) \\
`or\_70' & 0.06 (78\%) & 2.4 (100\%) \\
`or\_100' & 0.06 (73\%) & 2.4 (100\%) \\
`blasted' & 0.04 (80\%) & 2.4 (84\%) \\
`75' & 0.04 (92\%) & 2.0 (100\%) \\
`90' & 0.03 (16\%) & 12.4 (68\%) \\
\bottomrule
\end{tabular}
\caption{Root mean squared error (RMSE) of estimates of the natural logarithm of the number of satisfying solutions is shown.  The fraction of benchmarks within each category that each approximate counter was able to complete within the time limit of 5k seconds is shown in parentheses.}\label{tab:baselines_sat}
\end{table}

\begin{table*}[tb]
\centering
\begin{tabular}{@{}l ccc @{}}
\toprule
   & \multicolumn{3}{c}{\textbf{Runtimes By Percentile}}\\
  \cmidrule(lr){2-4}
  Category  & DSharp (0/70/85/100) & ApproxMC3 (0/70/100) & F2 (0/70/100) \\
\midrule
`or\_50'  & 0.0 / 0.8 / 12.4 / 48.1 & 0.1 / 336.6 / 5k & 0.2 / 4.0 / 89.9 \\
`or\_60'  & 0.0 / 0.3 / 2.1 / 79.1 & 0.1 / 276.6 / 5k & 0.2 / 5.0 / 353.2 \\
`or\_70'  & 0.0 / 0.7 / 3.6 / 46.6 & 0.1 / 748.3 / 5k & 0.2 / 11.9 / 491.3 \\
`or\_100'  & 0.0 / 0.3 / 4.8 / 54.2 & 0.1 / 1918.8 / 5k & 0.2 / 33.0 / 3021.1 \\
`blasted'  & 0.0 / 1.7 / 29.3 / 1390.8 & 0.0 / 952.6 / 5k & 0.0 / 742.3 / 5k \\
`75'  & 0.0 / 6.0 / 29.0 / 160.3 & 279.6 / 805.1 / 5k & 1.1 / 2.3 / 9.0 \\
`90'  & 0.0 / 1.8 / 16.7 / 479.9 & 326.3 / 5k / 5k & 1.1 / 5k / 5k \\
\bottomrule
\end{tabular}
\caption{Runtime percentiles (in seconds) are shown for DSharp, ApproxMC3, and F2.  Percentiles are computed separately for each category's training dataset.  In comparison, BPNN sequential runtime is nearly a constant and BPNN parallel runtime is limited by GPU memory.}\label{tab:runtimes_sat}
\end{table*}

\section{Stochastic Block Model Experiments} 
\label{SBM Appendix}

\paragraph{Stochastic Block Model Definition}
A $C$ class Stochastic Block Model (SBM) is a randomly generated graph with $N$ vertices, class assignment probabilities $p_{i}; i \in {1,\dots,C}$, where $\sum_{i=1}^C p_{i} = 1$, and edge probabilities $e_{ij}; i, j \in {1,\dots,C}$. Then, to generate the graph, we sample a class for each node, $c_{m}; m \in {1,\dots,N}$ in accordance with the class assignment probabilities. Then, we sample the edge set $E$ in the following manner: we take every pair of nodes $x_{m}, x_{n}; m, n \in {1,\dots,N}$ and with probability $e_{c_{m}, c_{n}}$ assign an edge between those nodes.

\paragraph{SBM Factor Graph Construction}
For a given SBM with $N$ nodes, $C$ classes, class assignment probabilities $p_{i}; i \in {1,\dots,C}$, sampled class assignments $c_{m}; m \in {1,\dots,N}$, edge probabilities $e_{ij}; i, j \in {1,..,C}$, and sampled edge set $E$, we have the following unary factor potentials $f_{i}(x_{m}); i \in {1,\dots,C}$ for every node $x_{m}; m \in {1,\dots,N}$:

$f_{i}(x_{m}) = p_{i}$

We can construct binary factor potentials $f_{ij}(x_{m}, x_{n}); i, j \in {1,..,C}$ between nodes $x_{m}, x_{n}; (m, n) \in E$ as:

$f_{ij}(x_{m}, x_{n}) = e_{c_{m}, c_{n}}$

and between nodes $x_{m}, x_{n}; (m, n) \notin E$ as:

$f_{ij}(x_{m}, x_{n}) = 1 - e_{c_{m}, c_{n}}$

Note that when we fix a variable to a specific value, we simply set all factor potentials involving that variable that do not agree with that value to zero. 

\paragraph{Marginal Calculation from Log Partitions}
Training a model to estimate partition functions with fixed variables is advantageous as we train the model to perform tasks that can directly be used to compute marginals, which are the probabilities that a node belongs to a specific class. This can be used to perform community detection or to quantify uncertainty and rare events in community membership. To see how we compute marginals with our experimental setup, take a two class SBM, select a node $x_m$, fix its value to class 0 to obtain log partition function $\ln(Z_{1})$ and then fix its value to class 1 to obtain log partition $\ln(Z_{2})$. Then, the log marginals $\ln(x_{m}^0)$ and $\ln(x_{m}^1)$ are simply:
    
$\ln(x_{m}^0) = \ln(Z_{1}) - \ln(Z_{1}+Z_{2})$
    
and
    
$\ln(x_{m}^1) = \ln(Z_{2}) - \ln(Z_{1}+Z_{2})$
    
where 
    
$\ln(Z_{1}+Z_{2})$
    
can be computed in a numerically stable fashion from $\ln(Z_{1})$ and $\ln(Z_{2})$ using the logsumexp trick.

\paragraph{Model and Training Details}
For baselines, we ran Belief Propagation to convergence with parallel updates and damping coefficient .5 as well as a Graph Isomorphism Network (GIN) with 30 layers and width 8. GIN is maximally discriminative among GNNs that consider 1-hop neighbors, which is computationally comparable to BPNN. In our evaluations, GIN performs comparably to more computationally expensive two hop GNNs on the related problem of SBM community detection \cite{Chen2019SupervisedCD}. We trained our GIN GNN architecture on the 5 class graph coloring community detection setting described in \cite{Chen2019SupervisedCD} and compared it to the performance of the two hop GNNs described there. Our GNN had 20 layers with a width of 8 and achieved a permutation invariant validation overlap score  of .166 when trained for the same number of iterations, nearly identical to the two hop GNN performance reported in \cite{Chen2019SupervisedCD}. Since one hop GNNs train significantly faster than two hop, we managed to obtain overlap scores as high as .185 when training for longer. In any case, our one hop GNN performs comparably with two hop GNN architectures on the related task of SBM community detection and thus, in addition to its convenience, makes for a strong baseline method. For all models, we trained for 300 epochs on 1 GPU with an Adam Optimizer (learning rate of 2e-4, batch size of 8) minimizing Mean Squared Error between the estimated log partition and true log partition.

\paragraph{Out of Distribution Generalization}

We test the capacity of our BPNN (with no double counting and an invariant BPNN-B layer) to generalize to out of distribution graphs compared to the GNN model, while comparing both against the BP benchmark. Since the factor graphs are fully connected, slight changes to the initial parameters can produce rather large differences in the graphs and their log partition function. In addition to perturbing the initial class probabilities and edge probabilities, we also test the ability of BPNN to generalize to larger graphs, which is a desirable property as the Junction Tree algorithm for exact inference becomes exponentially more expensive as the size of the graph grows due to the fully connected nature of SBM factor graphs. For each scenario, we generate five separate graphs and generate test examples as mentioned previously. We present our results in Table \ref{tab:bpnn_sbm_diff_dist_15}. We observe that BPNN performs the best of all three methods when class and edge probabilities are changed and generalizes better than GNN in these settings as well. Furthermore, when the size of graphs are increased, BPNN can outperform BP and GNN on graphs with as many as 20 nodes (a setting with over 80\% more edges than training) and generalizes significantly better than GNN. 

Since our SBM factor graphs are fully connected, adding $n$ times more nodes leads to a $O(n^2)$ increase in edges which may make it tougher for the model to generalize to larger and larger graphs. Using an auxiliary field approximation for SBM message passing, as described in \cite{PhysRevE.84.066106} can help generalization to larger graphs, as in this case the increase in edges will increase linearly with graph size, and this is something to investigate further. 

\begin{table}
    \begin{center}
    \begin{tabular}{c c c c c c} 
    \toprule
    \multicolumn{6}{c}{\textbf{Out of Distribution SBM RMSE}}  \\
    Nodes & Data Edge Probs & Data Class Probs & BP RMSE & GNN RMSE & BPNN RMSE \\
    \hline
    15 & (.93, .067) & (.6, .4) &  12.27 & 9.21 & 4.61\\ 
    \hline
    15 & (.93, .067) & (.8, .2) & 11.22 & 12.19 & 5.68 \\
    \hline
    15 & (.967, .033) & (.6, .4) & 16.77 & 12.62 & 4.92 \\
    \hline
    15 & (.9, .1) & (.8, .2) & 8.99 & 16.83 & 6.53 \\
    \hline
    15 & (.967, .13) & (.75, .25) & 9.54 & 12.77 & 6.64\\
    \hline
    15 & (.867, .033) & (.75, .25) & 11.55 & 9.17 & 4.14\\
    \hline
    16 & (.9375, .0625) & (.75, .25) & 13.88 & 15.07 & 7.08\\
    \hline
    17 & (.94, .06) & (.75, .25 & 15.92 & 17.89 & 8.43\\
    \hline
    18 & (.94, .06) & (.75, .25) & 15.81 & 20.90 & 10.50 \\
    \hline
    19 & (.95, .05) & (.75, .25) & 18.6 & 22.77 & 10.61 \\
    \hline
    20 & (.95, .05) & (.75, .25) & 19.37 & 28.31 & 15.23 \\
    \hline
    \end{tabular}
    \end{center}
    \label{SBM Out of Distribution Performance}
    \caption{RMSE of $\ln(Z)$ of BPNN against BP and GNN for SBM's generated from different distributions and larger graphs than the training or validation set. We see that BPNN outperforms both methods here across different edge probabilities, class probabilities, and on larger graphs. Furthermore, it generalizes better than GNN in all these settings.}
    \label{tab:bpnn_sbm_diff_dist_15}
\end{table}

\paragraph{Marginal Estimation}
We also compared BPNN to BP for marginal estimation, using the estimated log partition functions with single nodes set to a fixed value to calculate marginals for those nodes, as described above. Under the graph parameters used in these experiments, the marginals are usually extremely close to 1 and 0, but in such dense graphs, changes to the magnitude of these marginals can have large effects on the log partition function calculation. In some cases, BP computes the correct marginals under these conditions, but in some cases, it is off by 20-30 orders of magnitude on the smaller marginal. Such errors do not affect community recovery, however, when we care about very rare outcomes, they can have a big effect on quantifying uncertainty in community membership. On 15 node graphs, BPNN, by learning more accurate log partitions, is on average almost 5 orders of magnitude closer to the true marginals than BP but only an order more accurate than GNN. We see that on marginals, BPNN's overall performance and generalization ability relative to GNN is not as strong as it was with estimating partitions, likely because it is not specifically trained to estimate marginals, and estimating partitions and marginals, while quite related, are still different tasks. Training explicitly to estimate marginals, e.g. by correctly predicting the difference in partitions between graphs with one variable fixed to either value, may help performance and generalization ability of BPNN on marginals, and this is an area of further investigation.

\end{document}